\documentclass{article}
\usepackage{algorithm}

\usepackage{amsmath} 
\usepackage[preprint]{corl_2026} 

\usepackage{subcaption}
\usepackage{booktabs}
\usepackage{caption}
\usepackage{enumitem}
\usepackage{wrapfig}
\usepackage{comment}
\usepackage{multirow}
\usepackage{algpseudocode}
\usepackage{graphicx}
\usepackage{arydshln}
\title{SAFECAST: Robust Failure Detection for VLA Policies with Contrast-Set Training and Calibration}

\author{
  Harshitha Rajaprakash \hspace{3em} Aditeya Prajapati \hspace{3em} Rong Xue \vspace{4pt} \\ \vspace{2pt}
  \textbf{Abrar Anwar} \hspace{3em} \textbf{Jesse Thomason} \\
  University of Southern California,
  Thomas Lord Department of Computer Science \\
  \texttt{rajaprak@usc.edu}
}

\newcommand{\DpiTrain}{\mathcal{D}_{\pi}^{\mathrm{train}}}

\newcommand{\DprobeTrain}{\mathcal{D}_{\phi}^{\mathrm{train}}}
\newcommand{\DprobeCal}{\mathcal{D}_{\phi}^{\mathrm{cal}}}

\newcommand{\Dsrc}{\mathcal{D}_{\mathrm{src}}}
\newcommand{\DsrcTrain}{\mathcal{D}_{\phi,\mathrm{src}}^{\mathrm{train}}}
\newcommand{\DsrcCal}{\mathcal{D}_{\phi,\mathrm{src}}^{\mathrm{cal}}}

\newcommand{\Dcs}{\mathcal{D}_{\mathrm{CS}}}
\newcommand{\DcsVis}{\mathcal{D}_{\mathrm{CS}}^{\mathrm{vis}}}
\newcommand{\DcsLang}{\mathcal{D}_{\mathrm{CS}}^{\mathrm{lang}}}

\newcommand{\DcsTrain}{\mathcal{D}_{\phi,\mathrm{CS}}^{\mathrm{train}}}
\newcommand{\DcsCal}{\mathcal{D}_{\phi,\mathrm{CS}}^{\mathrm{cal}}}

\newcommand{\DcsVisTrain}{\mathcal{D}_{\phi,\mathrm{vis}}^{\mathrm{train}}}
\newcommand{\DcsLangTrain}{\mathcal{D}_{\phi,\mathrm{lang}}^{\mathrm{train}}}
\newcommand{\DcsVLTrain}{\mathcal{D}_{\phi,\mathrm{vis+lang}}^{\mathrm{train}}}

\newcommand{\DcsVisCal}{\mathcal{D}_{\phi,\mathrm{vis}}^{\mathrm{cal}}}
\newcommand{\DcsLangCal}{\mathcal{D}_{\phi,\mathrm{lang}}^{\mathrm{cal}}}
\newcommand{\DcsVLCal}{\mathcal{D}_{\phi,\mathrm{vis+lang}}^{\mathrm{cal}}}

\newcommand{\Daug}{\mathcal{D}_{\mathrm{aug}}}
\newcommand{\DaugTrain}{\mathcal{D}_{\phi,\mathrm{aug}}^{\mathrm{train}}}
\newcommand{\DaugCal}{\mathcal{D}_{\phi,\mathrm{aug}}^{\mathrm{cal}}}

\newcommand{\Deval}{\mathcal{D}_{\mathrm{eval}}}

\newcommand{\DlibSpatial}{\mathcal{D}_{\mathrm{LIBERO\text{-}Spatial}}}

\newcommand{\DdroidSpatial}{\mathcal{D}_{\mathrm{DROID}}}

\newcommand{\tauSrc}{\tau^{\mathrm{src}}}
\newcommand{\tauCS}{\tau^{\mathrm{CS}}}

\newcommand{\hiddenT}{h_t}

\newcommand{\riskT}{r_t}
\newcommand{\riskTraj}{r_{1:T}}

\newcommand{\probe}{f_{\phi}}

\newcommand{\thresholdT}{\delta_t}
\newcommand{\Scal}{S_{\mathrm{cal}}}
\newcommand{\alphacp}{\alpha}
\newcommand{\qalpha}{q_{\alpha}}

\newcommand{\Cone}{\textsc{SAFE}}
\newcommand{\Ctwo}{\textsc{SAFECAST}_\textsc{TrainAug}}
\newcommand{\Cthree}{\textsc{SAFECAST}_\textsc{CalAug}}
\newcommand{\Cfour}{\textsc{SAFECAST}}

\begin{document}
\maketitle


\begin{abstract}

Vision-language-action policies often fail under deployment-time distribution shifts such as clutter, distractor objects, lighting changes, novel objects, altered initial states, and reworded instructions.
Hidden-state-based risk probes combined with functional conformal prediction can detect rollout failures, but their reliability depends on calibration data matching deployment conditions.
We introduce SAFECAST, which leverages contrast set perturbations to improve hidden-state probe training and calibration for deployment time shift.
SAFECAST statistically significantly improves failure detection ROC-AUC scores over a state of the art baseline in both real-world DROID and LIBERO simulation experiments across multiple VLM backbones.
We further find that SAFECAST benefits most when both visual and language contrast set perturbations are used to augment data, and that with contrast set perturbations, sim-to-real calibration leads to better probes than using real rollout data only.


\end{abstract}


\section{Introduction}

Vision-language-action (VLA) policies enable robots to perform manipulation tasks from natural language instructions. 
Models such as OpenVLA~\cite{kim2024openvla} and $\pi_0$~\cite{black2024pi_0} exhibit promising generalization across tasks, objects, and environments.
However, 
a policy that succeeds in a clean training-like setting may fail when exposed to clutter, distractor objects, lighting variation, novel object appearances, changed robot initial states, or reworded instructions. 
These failures are concerning in safety-critical settings such as assistive robotics, human-robot interaction, and manipulation of fragile objects.

\begin{figure}[!t]
    \centering
    \begin{minipage}[b]{1\linewidth}
        \centering
        \includegraphics[width=\linewidth]{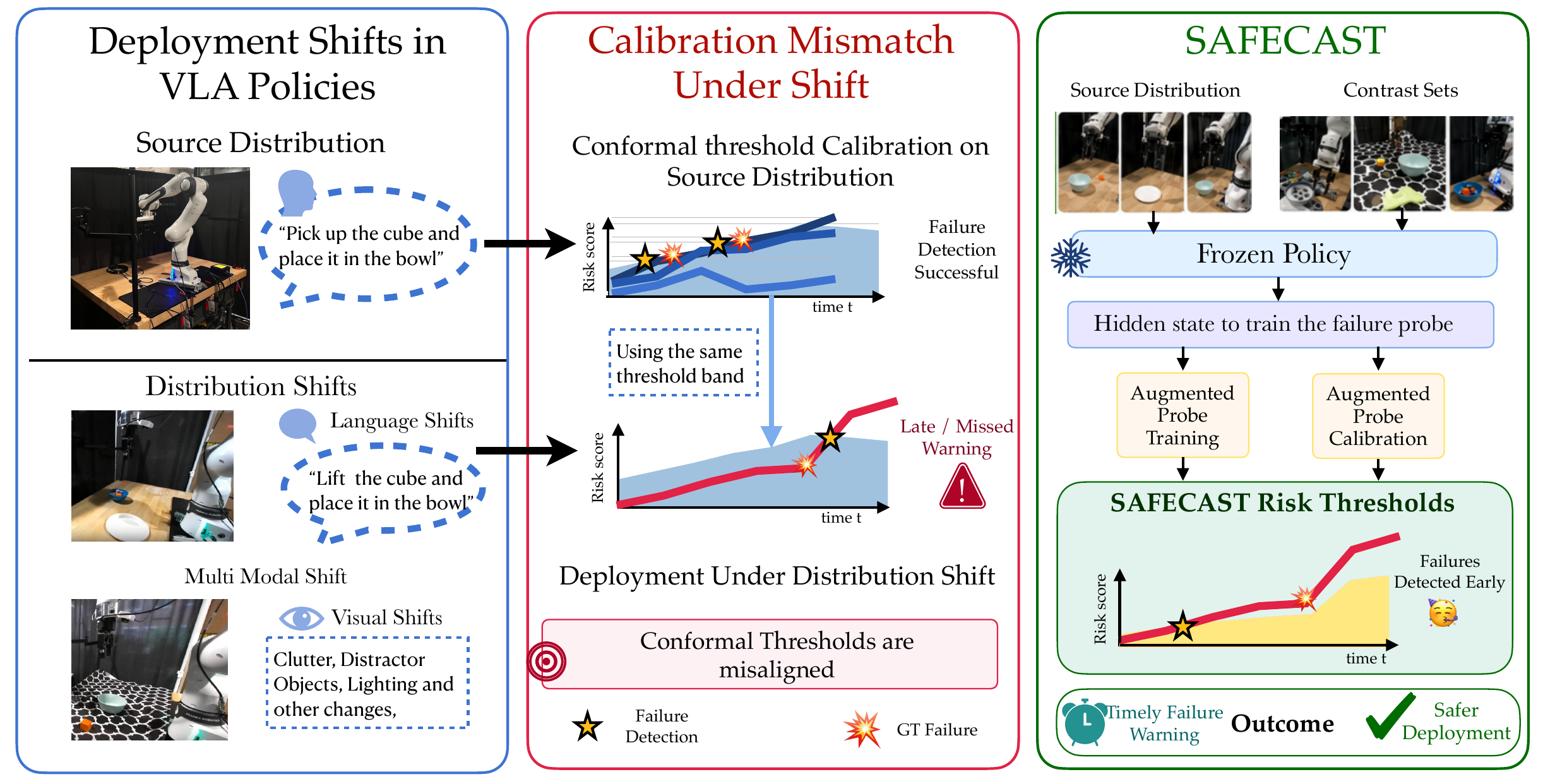}
        \vspace{-12pt}
    \end{minipage}
\caption{\textbf{SAFECAST.}
\textit{Left:} VLA policies trained on source-distribution rollouts can encounter visual, language, and multimodal deployment shifts during execution. 
\textit{Middle:} Hidden-state failure detectors calibrated only on source-distribution trajectories can become misaligned under deployment mismatch, leading to delayed or missed failure warnings. 
\textit{Right:} SAFECAST intervenes at probe training and conformal calibration time with contrast-set trajectories generated from perturbed in-distribution conditions, producing more reliable time-varying risk thresholds during deployment.
}
\label{fig:overview}
\end{figure}

A growing line of work studies runtime failure prediction for robotic policies. 
The existing \textbf{S}c\textbf{A}lable \textbf{F}ailure \textbf{E}stimation (SAFE) \cite{safe}-style methods use hidden states of VLA policies to predict failure risk and apply functional conformal prediction to obtain temporally calibrated risk thresholds, flagging unsafe rollouts before task failure occurs. 
However, their reliability depends on a key assumption: that the calibration trajectories resemble the trajectories encountered during deployment.
Deployment rollouts may differ due to visual shifts, language shifts, or their combination. 
Under such multimodal shifts, the distribution of hidden-state risk trajectories can change substantially, causing thresholds calibrated only on source-distribution rollouts to become unreliable. 

This work introduces SAFECAST (\textbf{S}c\textbf{A}lable \textbf{F}ailure \textbf{E}stimation with \textbf{C}ontrast-set \textbf{A}ugmentation for \textbf{S}afety \textbf{T}racking), a contrast-set-aware framework for robust hidden-state failure detection for VLAs.
SAFECAST incorporates visual and language contrast-set trajectories during probe training and conformal calibration (Figure~\ref{fig:overview}).
We find that \Cfour\ provides more reliable failure detection during distribution shifts than vanilla \Cone, that joint visual-language contrast sets improve robustness more than single-modality contrast sets, and that \Cfour\ can be trained in simulation and calibrated on physical rollouts to outperform more expensive real world probe training.

\section{Related Works}

Vision-language-action (VLA) policies demonstrate strong manipulation capabilities but remain sensitive to deployment-time distribution shifts.
Our work builds on runtime failure detection and conformal calibration for robot policies and studies whether contrast-set perturbations can improve failure detection robustness under  distribution shifts.

\textbf{Vision-Language-Action (VLA) Models} have demonstrated strong generalization across manipulation tasks by directly mapping visual observations and language instructions to robot actions~\cite{brohan2022rt1, brohan2023can, kim2024openvla, black2024pi0, physicalintelligence2025pi0.5, octo}.
These systems leverage large-scale robot demonstration data to enable broader capabilities without task-specific engineering.
However, prior work has shown that VLA performance degrades substantially under deployment-time perturbations such as clutter, lighting variation, and language perturbation~\cite{wilbert_colloseum, simpler_env, anwar2024contrast, anwar2025efficient, kress2024robot, parekh2024investigating,xie2024decomposing,taxonomy2025arxiv,mitra2025roboticsteering}.
While benchmark performance continues to improve through scaling~\cite{nasiriany2024robocasa, liu2023libero, fei25libero, yu2019meta}, reliable deployment remains challenging due to sensitivity to out-of-distribution conditions.
Our work builds on this setting and focuses on improving runtime reliability under deployment shift.

\textbf{Runtime Failure Detection for Robot Policies} is critical for safely deploying learned robot policies in real-world environments, where even small execution errors can lead to unsafe or irreversible outcomes~\cite{sinha2022system, natarajan2023fault, rahman2021runtime}.
Prior work has studied failure detection through out-of-distribution monitoring~\cite{xu2025faildetect, liu2024fleet, sinha2024anomaly, wong2022error, majumdar2025predictive, agia2025stac} and supervised failure prediction~\cite{liu2024runtime, gokmen2023asking, xie2022askhelp, sinha2023fallback, farid2022failure, ablett2020fighting}.
Several methods also estimate model uncertainty as a proxy for failure.
In language models, token-level uncertainty and semantic disagreement across sampled generations have been used to detect hallucinations and unreliable outputs~\cite{huang2023look, kuhn2023semantic, shorinwa2024survey}, and have been adapted as baselines for VLA failure detection~\cite{safe}.
Other approaches use large vision-language models~\cite{duan2024aha, du2023vlm_success} or reward models~\cite{liang2026robometer} for success or failure detection.
While these methods provide useful runtime monitoring signals, many require multiple action samples, additional model queries, or expensive VLM inference.
We build on state of the art work on multitask failure detection for vision-language-action policies that probes hidden policy representations and calibrating time-varying failure thresholds~\cite{safe}.
Internal policy representations contain information predictive of downstream failures.
However, existing failure detectors are generally calibrated under source-distribution conditions.
We directly study how to improve their reliability when deployment conditions differ substantially from calibration data.

\textbf{Conformal Calibration Under Deployment Shift.}
Conformal prediction provides calibrated decision thresholds~\cite{vovk2005algorithmic, angelopoulos2021gentle}, and functional conformal prediction extends these ideas to time-varying signals~\cite{diquigiovanni2024band}.
Recent robot failure detection methods use conformal prediction to set time-varying thresholds for sequential failure scores~\cite{xu2025faildetect, safe, farid2022failure}.
However, these guarantees rely on calibration rollouts being representative of deployment rollouts, an assumption frequently violated in NLP~\cite{devic2025calibration} and robotics.
Contrast sets have been used to expose model brittleness through targeted perturbations in NLP, vision-language reasoning, and robotics~\cite{gardner2020evaluating, anwar2024contrast}.
Our work uses contrast-set perturbations to study and improve calibration robustness, finding that perturbation-aware probe training and calibration can strengthen VLA failure detection under distribution shifts.

\section{Background and Task Definiton}
\label{sec:task}





Our method builds on \textbf{S}c\textbf{A}lable \textbf{F}ailure \textbf{E}stimation (SAFE)~\cite{safe}, a hidden-state risk probing framework for runtime failure detection in vision-language-action (VLA) policies. 
SAFE takes as input the hidden state of a frozen VLA policy at timestep $t$ together with a conformal significance level $\alpha$, corresponding to different operational risk tolerances during deployment, and predicts whether the ongoing rollout is likely to fail. 
The probe is trained on VLA rollouts labeled with success or failure to predict, at every timestep, whether failure will occur.
The probe's temporally-dependent threshold for runtime failure detection is calibrated with functional conformal prediction. 


Our proposed intervention targets the data distributions used for probe training and conformal calibration. 
Inspired by contrast sets~\cite{anwar2024contrast}, we augment these datasets with perturbed rollout trajectories constructed through visual and language perturbations that require minimal experimenter overhead. 
We denote this framework as SAFECAST (\textbf{S}c\textbf{A}lable \textbf{F}ailure \textbf{E}stimation with \textbf{C}ontrast-\textbf{S}et \textbf{A}ugmentation for \textbf{S}afety \textbf{T}racking).


\paragraph{Vision-Language-Action (VLA) Policies} map visual observations and language instructions to robot actions.
Let $\pi$ denote a frozen VLA policy.
At timestep $t$, the policy receives a visual observation $o_t$ and language instruction $l$, and predicts an action
$
a_t \sim \pi(\cdot \mid o_t, l).
$
Executing $\pi$ over a horizon of length $T$ produces a rollout
$
\tau = \{(o_t, l, a_t)\}_{t=1}^{T},
$
where $o_t$ depends on previous actions and environment state while $l$ remains fixed throughout the rollout.
Let $\DpiTrain$ denote the data used to train and finetune the VLA policy $\pi$. 
We fix $\pi$ after this training; the VLA policy is not updated during failure-probe training, conformal calibration, or evaluation.



\paragraph{Functional Conformal Prediction for VLAs.}

SAFE~\cite{safe} probes VLA hidden states to predict whether an ongoing rollout will eventually fail. 
Given a rollout $\tau$, the frozen VLA policy produces a sequence of hidden states during execution. 
For each timestep $t$, let $\hiddenT$ denote the pre-final-layer hidden representation extracted from $\pi$. 
A lightweight MLP probe $\probe$ maps this hidden representation to a scalar failure score,
$\riskT = \probe(\hiddenT).$ The score $\riskT$ is interpreted as the estimated risk that the rollout will eventually fail, conditioned on the current hidden state.

For a rollout of length $T$, applying the probe at every timestep produces a temporally aligned risk score sequence
${
\riskTraj = (r_1, r_2, \dots, r_T).
}$
This risk sequence is used for both conformal calibration and test-time failure detection.

After probe training, functional conformal prediction calibrates the time-varying failure threshold $\thresholdT$ using calibration rollouts $\tau_j^{\mathrm{cal}}$. 
For each calibration rollout, the trained probe produces a trajectory of risk scores $\riskTraj^{(j)}$, from which functional conformal prediction computes nonconformity scores, $\Scal$ over successful calibration trajectories to construct a one-sided time-varying threshold $\thresholdT = \mu_t + \qalpha$, and ${\alphacp \in (0,1)}$ controls the deployment-time operational risk tolerance. 

Lower $\alpha$ values yield higher risk thresholds leading to lower false positives and higher $\alpha$ results in a lower threshold which results in a higher false positive rate.
During deployment, the detector flags a rollout as headed for failure whenever $\riskT > \thresholdT$.

\paragraph{Problem Definition: Failure Detection Under Distribution Shift.}

This work studies reliable rollout-level failure detection for frozen VLA policies under deployment-time distribution shift.
Let $\Dsrc$ denote the source-distribution rollout pool from which probe training and calibration data are sampled. 
Let $\Deval$ denote the evaluation distribution encountered during deployment. 
The evaluation distribution may differ from $\Dsrc$ due to visual, language or multimodal their combination.


\section{SAFECAST}

We introduce SAFECAST for robust failure detection using hidden-state probes under multimodal deployment shift using contrast set training and calibration.
Given VLA rollouts from the source distribution $\Dsrc$ and a contrast-set pool $\Dcs$, hidden states are extracted from the frozen VLA policy, a lightweight risk probe is trained to predict timestep-wise failure scores, and functional conformal prediction is used to calibrate time-varying failure thresholds. 
The key question is whether SAFECAST training and calibration improve failure detection on the evaluation distribution $\Deval$.

\paragraph{Contrast Sets for Failure Probing.}
Recall that $\Dsrc$ denotes the source rollout distribution from which the probe training set $\DprobeTrain$ and calibration set $\DprobeCal$ are drawn. 
A source-distribution rollout is denoted as $\tauSrc = \{(o_t,l,a_t)\}_{t=1}^{T}$,
where $o_t$ is the visual observation, $l$ is the language instruction, and $a_t$ is the action produced by the frozen VLA policy.

SAFECAST constructs a contrast-set rollout pool $\Dcs$ by applying controlled perturbation operators to the visual and/or language inputs of source-distribution rollouts and then re-executing the frozen VLA policy under the perturbed condition. 
A contrast-set rollout is denoted as $\tauCS = \{(\tilde{o}_t,\tilde{l},\tilde{a}_t)\}_{t=1}^{T},$
where $\tilde{o}_t$ is a visually perturbed observation, $\tilde{l}$ is a perturbed or paraphrased language instruction, and $\tilde{a}_t$ is the action produced by re-executing the policy with these perturbed inputs. 
Because the policy is re-executed, $\tauCS$ is not treated as a relabeled or counterfactual copy of $\tauSrc$; the resulting action sequence $\tilde{a}_{1:T}$ and hidden-state trajectory may differ from the source rollout. 
Thus, we do not assume trajectory-level exchangeability between $\Dsrc$ and $\Dcs$. 
Instead, contrast sets are used to construct training and calibration distributions whose risk trajectories better approximate deployment shifts expected in $\Deval$.

We use this family of contrast-set perturbations, $\Dcs = \DcsVis \cup \DcsLang,$
where $\DcsVis$ contains visual contrast-set rollouts, and $\DcsLang$ contains language contrast-set rollouts. 
The contrast-set-augmented rollout pool is $\Daug = \Dsrc \cup \Dcs.$
Depending on the configuration, probe training and/or conformal calibration draw data from $\Daug$ rather than only from $\Dsrc$.

\textbf{Visual contrast sets.}
Visual contrast sets perturb the observation stream while keeping the language instruction fixed. 
In LIBERO, we construct visual contrast sets by adding distractor objects to LIBERO-Spatial scenes. 
Distractors are selected so that they do not change the intended task, are not referenced in the language instruction, and do not obstruct the nominal source-distribution rollout. 
Thus, the target object, action, and goal remain unchanged, while the visual context differs from the source distribution. 
In real-world DROID experiments, visual perturbations include distractor objects, cluttered backgrounds, novel objects, and altered object configurations.
Additionally, DROID visual perturbations result in novel rollouts. For more details on the number of rollouts used per perturbation, please refer to Appendix~\ref{abb:cscreation}

\textbf{Language contrast sets.}
Language contrast sets perturb the instruction while keeping the visual scene fixed. 
In LIBERO, we generate three paraphrases for each LIBERO-Spatial task, by prompting the ChatGPT 5.1 Instant model. For more details regarding the paraphrases and prompt used to generate them, refer to Appendix~\ref{abb:cscreation}. These are designed to  preserve the intended target object, action, and goal location. 
To reduce rollout collection cost, each paraphrase is evaluated on three episodes per task rather than all 50 source episodes. 
These perturbations test whether the failure detector remains reliable when task semantics are preserved but the linguistic form changes. 
In real-world DROID experiments, language perturbations include paraphrased instructions and negated or distractor phrasing when applicable. These were generated by giving a prompt to ChatGPT 5.1. 

\textbf{Joint visual-language contrast sets.}
Joint visual-language contrast sets perturb both the observation stream and the language instruction. 
These rollouts combine visual perturbations with paraphrased instructions and therefore induce multimodal deployment shifts. 
They are intended to approximate deployment conditions that may violate the exchangeability assumptions underlying source-distribution conformal calibration. Examples of these are primarily in the real setting.

Candidate training rollouts are filtered using an active rejection procedure based on Dynamic Time Warping (DTW). 
Candidates whose trajectories are too similar to already-selected rollouts are rejected using a fixed DTW distance threshold, reducing near-duplicates in $\DcsTrain$. For more details, refer to \ref{dtw}
The augmented probe-training set is then
$\DaugTrain 
$.
For conformal calibration, the contrast-set calibration pool is
$\DcsCal 
$
and the augmented calibration set is
$\DaugCal 
$
Calibration subsets are sampled with controlled success and failure counts where possible so that threshold estimation is not dominated by a single outcome class. 
Appendix~\ref{app:training_vs_calibration} reports the number of rollouts collected for each perturbation type and split.

\paragraph{Contrast Set-Augmented Probe Training and Calibration.}
SAFE probe training and calibration sample rollouts from $\DsrcTrain$ and $\DsrcCal$, respectively, while SAFECAST instead uses the augmented rollout datasets $\DaugTrain$ and $\DaugCal$ containing contrast-set trajectories.

\section{Experiments and Results}

\begin{table}[t]
  \centering
\caption{
    Failure detection performance across simulation and real-world settings comparing SAFE and SAFECAST variants.
    Contrast set-aware probe training and calibration consistently improve failure detection over the SAFE method across both F1 and ROC-AUC.
    F1 measures the balance between precision and recall for rollout-level failure detection, while ROC-AUC measures how well the failure scores separate successful and failed rollouts.
    Averaged over $\alpha \in \{0.1, 0.2, \ldots, 0.9\}$ and 30 seeds).
    \textbf{Bold} indicates maximum.
    $^*$ indicates statistical significance, tested for ROC-AUC across paired performance at each $\alpha$ threshold.
    Exact $p$ values are reported Appendix~\ref{app:stats}.
    }
  \label{tab:alpha-marginalized-f1-roc-combined-sim-real}
  \small
  \setlength{\tabcolsep}{4pt}
  \renewcommand{\arraystretch}{1.08}
  \resizebox{\linewidth}{!}{
  \begin{tabular}{lcccccccc}
    \toprule
    & \multicolumn{4}{c}{F1} & \multicolumn{4}{c}{ROC-AUC} \\
    \cmidrule(lr){2-5}\cmidrule(lr){6-9}
    Method
    & Sim-$\pi_0$
    & Sim-OpenVLA
    & Real-$\pi_0$
    & Real-$\pi_0$-FAST
    & Sim-$\pi_0$
    & Sim-OpenVLA
    & Real-$\pi_0$
    & Real-$\pi_0$-FAST \\
    \midrule
    \textit{Always guess failure}
    & 0.5266
    & 0.8314
    & \textbf{0.7322}
    & 0.7654
    & 0.0000\phantom{$^*$}
    & 0.0000\phantom{$^*$}
    & 0.0000\phantom{$^*$}
    & 0.0000\phantom{$^*$} \\
    \midrule
    $\Cone$
    & 0.7132
    & 0.8589
    & 0.4085
    & 0.7475
    & 0.3267\phantom{$^*$}
    & 0.5937\phantom{$^*$}
    & 0.2626\phantom{$^*$}
    & 0.5501\phantom{$^*$} \\
    $\Ctwo$
    & 0.7516
    & 0.8687
    & 0.5838
    & \textbf{0.8109}
    & 0.4278$^*$
    & 0.6514\phantom{$^*$}
    & 0.2649\phantom{$^*$}
    & 0.6621$^*$ \\
    $\Cthree$
    & 0.7373
    & 0.8628
    & 0.4480
    & 0.7534
    & 0.4382$^*$
    & 0.6782$^*$
    & 0.3415$^*$
    & 0.6578$^*$ \\
    $\textbf{SAFECAST}$
    & \textbf{0.7528}
    & \textbf{0.8728}
    & 0.5412
    & 0.8053
    & \textbf{0.4469}$^*$
    & \textbf{0.8014}\phantom{$^*$}
    & \textbf{0.3807}$^*$
    & \textbf{0.6664}$^*$ \\
    \bottomrule
  \end{tabular}
  }
\end{table}

The main findings of our experiments are that (1) \Cfour\ provides more reliable failure detection during distribution shifts than vanilla \Cone; (2) joint visual-language contrast sets improve robustness more than single-modality contrast sets; and (3) \Cfour\ can be trained in simulation and calibrated on physical rollouts to outperform more expensive real world probe training.

\subsection{Experimental Setup}

We conduct experiments comparing \Cone\ and \Cfour\ on both a physical robot and in a simulation environment.
To isolate the roles of contrast-set probe training and contrast-set conformal calibration proposed as part of \Cfour, we evaluate four configurations (Table 1).
Additional details on the specific conditions and examples of DROID tasks, and on LIBERO perturbation procedures, policy performance, dataset filtering, and the simulation setup are included in Appendix~\ref{app:experiment_setup}.

We evaluate SAFECAST on a Franka/DROID setup using $\pi_0$ and $\pi_0$-FAST. 
For $\pi_0$, the open-source DROID checkpoint is LoRA-finetuned on ${\mathcal{D}_{\pi}^{\mathrm{train},\,\mathrm{real}}}$ consisting of 20 teleoperated spatial pick-and-place demonstrations, as the zero-shot checkpoint achieved zero successful rollouts in preliminary testing.
The $\pi_0$-FAST checkpoint is evaluated without additional finetuning since it achieved nonzero task success in the same setup. 
Evaluation is performed on deployment rollouts $\tau^{\mathrm{eval}} \in \Deval^{\mathrm{real}}$ containing stronger shifts including novel objects, new tasks, clutter, lighting variation, and altered robot initial states. 

We evaluate OpenVLA and $\pi_0$ on LIBERO-Spatial and LIBERO-Plus \cite{fei25libero}. 
Unlike SAFE~\cite{safe}, which splits LIBERO-Spatial into seen and unseen tasks from the same distribution, we separate source, calibration, and deployment distributions to study robustness under controlled deployment shift.

\textbf{Metrics.}
Following prior work in uncertainty estimation and VLA failure detection~\cite{kuhn2023semantic, huang2023look, safe}, we evaluate failure detectors using F1-score and ROC-AUC across sweeps of the conformal significance level $\alpha$. 
Each value of $\alpha$ induces a different calibrated intervention threshold, corresponding to a different deployment-time risk tolerance. 
Lower $\alpha$ values result in lower false positive rates (FPR) of flagging failure, while higher $\alpha$ values yield higher FPR, with more trajectories being flagged as headed for failure. 
We report $\alpha$-marginalized F1 by averaging over the evaluated $\alpha$ sweep, capturing robustness across risk tolerances rather than performance at a single selected threshold.
We compute TPR and FPR at each $\alpha$ and report ROC-AUC as the area under the resulting TPR--FPR curve. 

\begin{figure}[t]
    \centering
        \centering
        \includegraphics[width=.8\linewidth]{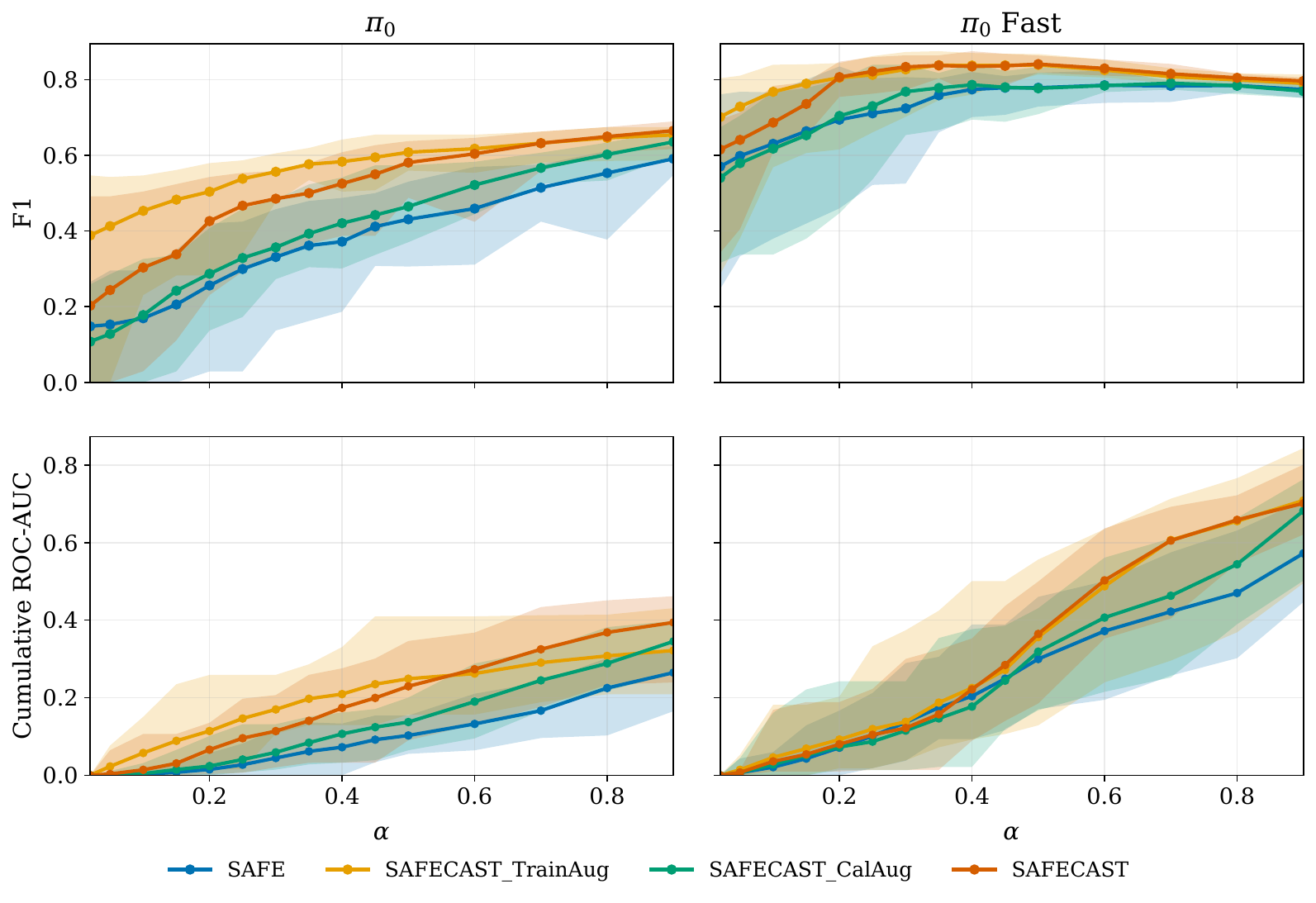}
        
\caption{Real-world failure detection across varying $\alpha$ on the DROID setup. 
Top: F1-score across calibrated intervention thresholds. Bottom: ROC-AUC measuring separation between successful and failed rollouts. 
Lower $\alpha$ values correspond to more conservative deployment regimes with stricter intervention thresholds. 
Across both $\pi_0$ and $\pi_0$-FAST, SAFECAST probes consistently improve failure detection performance over SAFE, with the strongest gains appearing when both augmented training and calibration are combined in SAFECAST. Averaged over 30 seeds.
}
        
        \label{fig:droid_seed_sweep_refactor}

    \hfill
    \vspace{0.5em}

\end{figure}

\subsection{SAFECAST improves failure detection during distribution shifts.}
\label{sec:main_results}

In the real world DROID experiments, SAFECAST substantially improves deployment robustness for both $\pi_0$ and $\pi_0$-FAST (Table~\ref{tab:alpha-marginalized-f1-roc-combined-sim-real}) across a broad range of risk tolerances (Figure~\ref{fig:droid_seed_sweep_refactor}).
Interestingly, the relative benefits of augmented probe training and augmented calibration differ across policies. 
For $\pi_0$, augmented probe training provides the largest gains, suggesting that exposure to perturbed rollout trajectories during representation learning substantially improves downstream failure separability. 
In contrast, $\pi_0$-FAST benefits more strongly from SAFECAST calibration, possibly indicating that calibration mismatch between source-distribution trajectories and deployment conditions becomes a larger bottleneck for stronger pretrained policies. 
Across both policies, SAFECAST either achieves the best overall performance or closely matches the strongest individual augmentation strategy and suggests that jointly incorporating perturbation-aware probe training and calibration produces the most consistently robust deployment-time failure detection behavior.

\begin{figure}
    \centering
    \includegraphics[width=.8\linewidth]{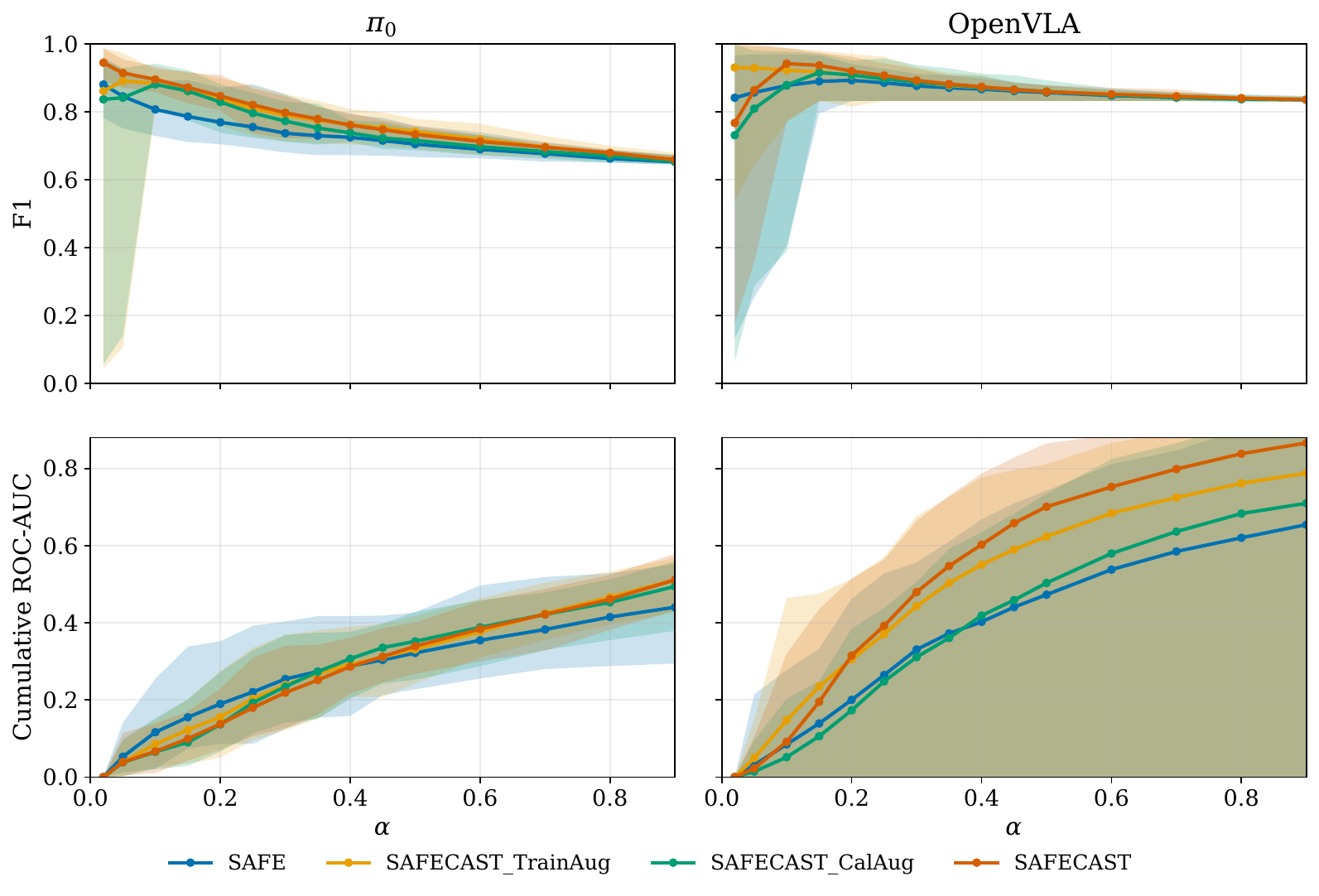}
\caption{Simulation failure detection across varying $\alpha$ on LIBERO. 
Top: F1 across the $\alpha$ values. Bottom: cumulative ROC-AUC across the $\alpha$ values
Across both $\pi_0$ and OpenVLA, contrast-set-aware probes consistently improve failure detection performance over SAFE, with SAFECAST and augmented probe training producing the strongest overall improvements. 
}
    
    \label{fig:sim_refactor}
\end{figure}

We find these results hold similarly in simulation. 
In LIBERO, we show a similar trend in Table~\ref{tab:alpha-marginalized-f1-roc-combined-sim-real} and Figure~\ref{fig:sim_refactor}. 
Across both OpenVLA and $\pi_0$, contrast-set-aware configurations consistently improve failure detection robustness over SAFE. 
As in the real-world setting, the strongest augmentation strategy varies across policies: OpenVLA benefits most from augmented probe training, while $\pi_0$ achieves the strongest overall robustness under full SAFECAST training and calibration. 
Together, these results support that SAFECAST strengthens the reliability of failure detection during diverse deployment conditions across various risk tolerances.


\subsection{Visual + Language Contrast Sets Improve Failure Detection Robustness}

\begin{figure}[t]
    \centering

    \begin{minipage}[c]{0.60\linewidth}
        \centering
        \includegraphics[width=\linewidth]{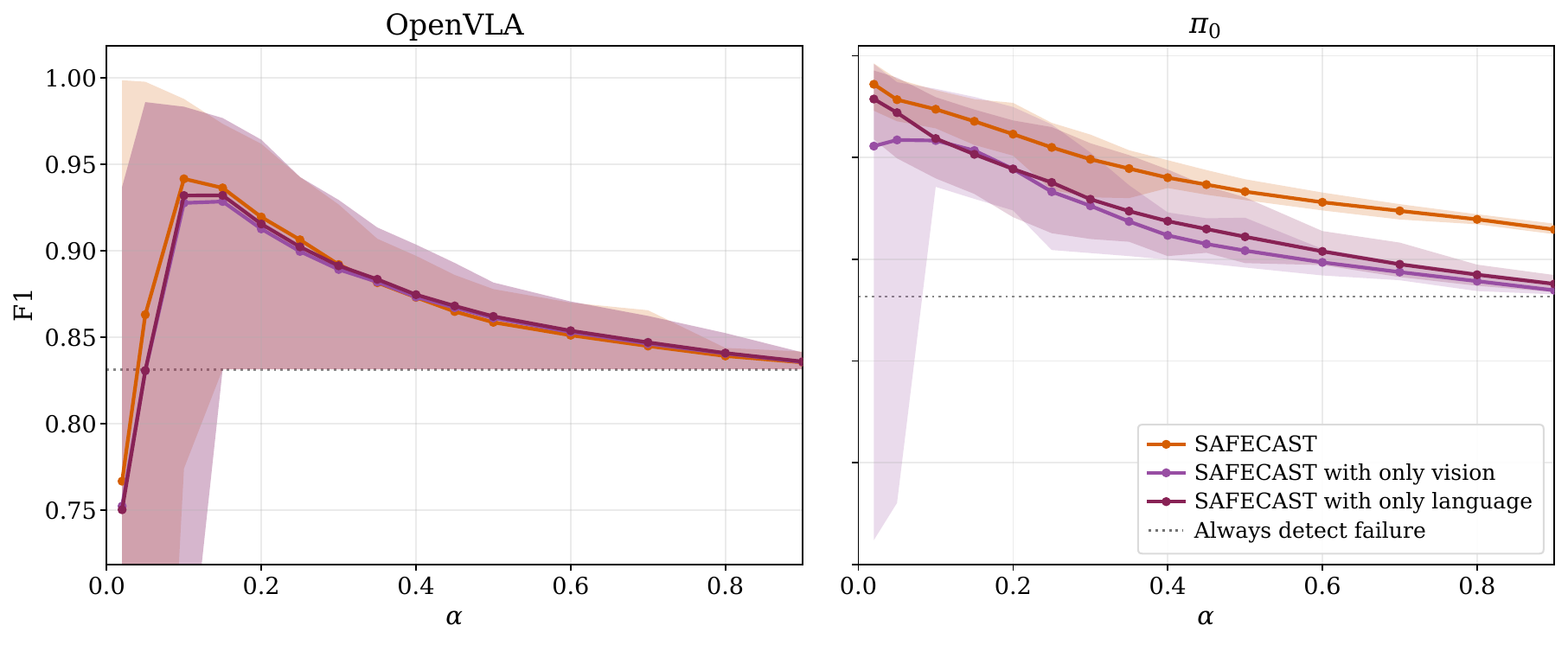}
    \end{minipage}
    \hfill
    \begin{minipage}[c]{0.36\linewidth}
        \centering
        \small
        \textbf{Mean F1 by Modality}\\[1mm]
        \begin{tabular}{lcc}
            \toprule
            Contrast Sets & OpenVLA $\uparrow$ & $\pi_0$ $\uparrow$ \\
            \midrule
            Visual only       & 0.871 & 0.649 \\
            Language only     & 0.872 & 0.664 \\
            Visual + Lang.    & \textbf{0.873} & \textbf{0.753} \\
            \bottomrule
        \end{tabular}
    \end{minipage}

    \vspace{-1mm}
    \caption{
    Effect of contrast-set modality on simulation failure detection. 
    Left: F1 across conformal significance levels $\alpha$ for OpenVLA and $\pi_0$. 
    Right: peak F1 for each contrast-set modality. 
    Joint visual-language contrast sets perform best for both policies.
    }
    \label{fig:simmodalityablations}
\end{figure}
Figure~\ref{fig:simmodalityablations} ablates visual-only, language-only, and joint visual-language contrast sets under the best-performing SAFECAST configuration from the simulation results in Section~\ref{sec:main_results}. 
Across both OpenVLA and $\pi_0$, jointly incorporating visual and language perturbations produces the strongest OOD failure detection robustness. 
For example, on OpenVLA, multimodal contrast sets improve F1 from $0.922$ under visual-only augmentation to $0.959$. 
While both visual-only and language-only contrast sets improve robustness, the best performance is consistently achieved when both modalities are included together, suggesting that deployment-time calibration benefits from covering the joint perturbation space expected during execution rather than modeling visual or language shifts independently. Please refer to \ref{app:modality} for ROC-AUC analysis.

\subsection{SAFECAST Probe Training in Simulation Calibrates DROID Failure Detection}

\begin{wrapfigure}{r}{0.5\linewidth}
\vspace{-4mm}
    \centering
    \includegraphics[width=\linewidth]{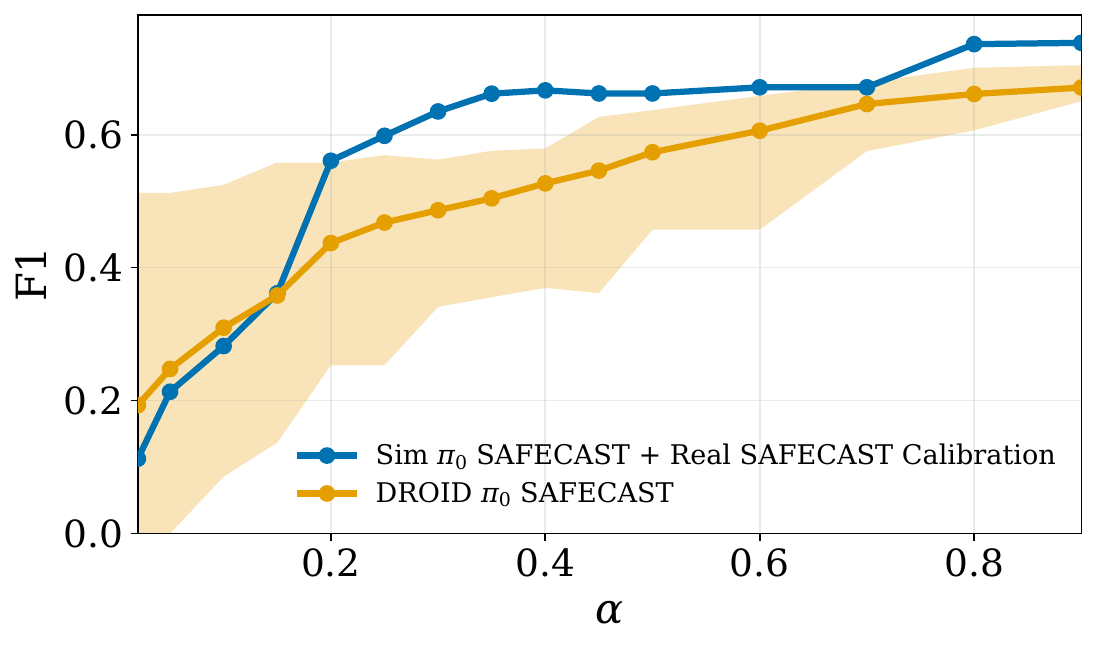}
    \vspace{-18pt}
    \caption{Sim-to-real failure detection robustness across varying $\alpha$. 
    SAFECAST enables sim-to-real transfer even under deployment mismatch. }
    \label{fig:sim2realh4}
\end{wrapfigure}

Since it is easier to collect large amounts of data in simulation, we train failure probes on trajectories generated in LIBERO-Spatial and then adapt calibration using a smaller real-world contrast-set dataset. 
Specifically, the probe is trained entirely in simulation while functional conformal calibration is performed using real-world contrast-set trajectories collected on the DROID setup. 
This sim-training to real-contrast-set-calibration  improves failure detection robustness compared to only training and calibrating a probe only on the smaller set of real-world data (Figure~\ref{fig:sim2realh4}). For additional analyses of ROC-AUC, please refer to \ref{app:sim2real}





\section{Conclusion}
\label{sec:conclusion}

We introduced SAFECAST, a framework for improving hidden-state failure detection in vision-language-action policies through contrast-set-aware probe training and conformal calibration.
SAFECAST addresses a key limitation of source-distribution failure calibration: deployment rollouts may induce visual, language, multimodal, or sim-to-real shifts that change the hidden-state risk trajectories used for failure detection.
By incorporating contrast-set trajectories during probe training and/or calibration, SAFECAST improves alignment between calibration and deployment risk distributions.

Across LIBERO simulation, real-world DROID/Franka experiments, and sim-to-real transfer, SAFECAST improves runtime failure detection under deployment shift.
The results show that contrast-set-aware configurations improve both F1 and ROC-AUC across conformal significance levels $\alpha$.
This supports the central claim that robust failure detection should be evaluated not only at a single operating point, but across deployment risk tolerances.

\textbf{Limitations.}
SAFECAST depends on collecting contrast-set trajectories that are representative of likely deployment shifts.
If the robot encounters perturbations outside the contrast-set distribution, failure detection performance may still degrade.
Additionally, contrast-set collection introduces extra rollout cost, and our current perturbation families cover only a subset of possible visual, language, and task-level changes.
The method also improves calibration alignment empirically, but does not restore formal conformal exchangeability guarantees under arbitrary deployment shift.

Future work could study adaptive contrast-set construction, where new perturbations are generated online based on detector uncertainty or observed failure modes.
Another promising direction is integrating SAFECAST with human-in-the-loop learning.
Calibrated risk estimates could trigger human intervention during execution, allowing supervisors to stop unsafe rollouts, provide corrective actions, or teleoperate through difficult states.
These interventions could then be incorporated back into the policy training and failure-calibration datasets, moving from passive failure detection toward interactive policy improvement.

\clearpage

\section*{Acknowledgments}
This work was funded in part by a DARPA ARC Safe and Assured Foundation Robots for Open Environments (SAFRON) grant (Award HR0011-25-3-0154).
The authors would like to thank grant Co-PIs Souti Chattopadhyay and William G.J. Halfond for brainstorming discussions.

\bibliography{references}

\newpage

\appendix
\pagenumbering{gobble}
\section{Contrast-Set Construction}
\label{app:contrast_construction}
\label{abb:cscreation}

This appendix provides additional details on the construction of source-distribution and contrast-set rollout pools used for SAFECAST training and calibration. All contrast-set rollouts are generated by perturbing the visual and/or language inputs of source-distribution conditions and re-executing the frozen VLA policy under the perturbed condition.

\subsection{Rollout Split Statistics}
\label{app:rollout_split_stats}

\begin{table}[htbp]
\centering
\small
\caption{Probe training and calibration rollout statistics for source-distribution and contrast-set datasets used in SAFECAST evaluation. Each entry reports the number of labeled rollout pairs $(\tau,y)$ in the source-distribution split and the contrast-set split. Source denotes $\Dsrc$, and CS denotes $\Dcs$.}
\label{tab:dataset_summary}
\begin{tabular}{llccccc}
\toprule
Setting & Policy & Split & $\Dsrc$ Succ. & $\Dsrc$ Fail. & $\Dcs$ Succ. & $\Dcs$ Fail. \\
\midrule
$\DlibSpatial$ & $\pi_0$ & Train & 436 & 16 & 72 & 15 \\
$\DlibSpatial$ & $\pi_0$ & Cal.  & 45  & 5  & 53 & 37 \\
$\DlibSpatial$ & OpenVLA & Train & 369 & 81 & 144 & 112 \\
$\DlibSpatial$ & OpenVLA & Cal.  & 42  & 8  & 23  & 94 \\
\midrule
$\DdroidSpatial$ & $\pi_0$ & Train & 17 & 27 & 6  & 26 \\
$\DdroidSpatial$ & $\pi_0$ & Cal.  & 14 & 28 & 9  & 21 \\
$\DdroidSpatial$ & $\pi_0$-FAST & Train & 25 & 22 & 13 & 18 \\
$\DdroidSpatial$ & $\pi_0$-FAST & Cal.  & 19 & 19 & 12 & 20 \\
\bottomrule
\end{tabular}
\end{table}

\subsection{Perturbation Types}
\label{app:perturbation_types}

\begin{table}[htbp]
\centering
\small
\caption{Contrast-set perturbations used to construct deployment-shift training and calibration rollouts. Counts report the number of contrast-set rollouts available for each policy and split.}
\label{tab:contrast_sets}
\begin{tabular}{llcccc}
\toprule
Setting & Perturbation Type & OpenVLA Train & OpenVLA Cal. & $\pi_0$ Train & $\pi_0$ Cal. \\
\midrule
Simulation & Visual distractors & 133 & 107 & 87  & 45 \\
Simulation & Language paraphrases & 133 & 25  & 103 & 45 \\
\midrule
Real robot & Visual distractors/clutter & 15 & 8  & 15 & 8  \\
Real robot & Language paraphrases/negations & 31 & 30 & 31 & 30 \\
Real robot & Visual + language & 32 & 30 & 32 & 30 \\
Real robot & Novel objects/tasks & 2  & 8  & 2  & 8  \\
\bottomrule
\end{tabular}
\end{table}

\paragraph{Simulation visual perturbations.}
In LIBERO-Spatial, visual contrast sets are generated by adding distractor objects to the scene. Distractors are selected so that they do not change the intended task, are not referenced in the language instruction, and do not obstruct the nominal source-distribution robot trajectory. Thus, the target object, action, and goal remain unchanged, while the visual context differs from the source distribution.

\paragraph{Simulation language perturbations.}
For each LIBERO-Spatial task, we generate three semantically equivalent instruction variants that preserve the intended target object, action, and goal location. To reduce rollout collection cost, each paraphrase is evaluated on three episodes per task rather than all 50 source episodes.

\paragraph{Language paraphrase prompt.}
The paraphrases are generated with the following prompt template:
\begin{quote}
\textit{Rewrite the following robot manipulation instruction in three semantically equivalent ways. Preserve the target object, action, and goal location. Do not introduce new objects, new goals, or ambiguous references. Instruction: ``[ORIGINAL INSTRUCTION]''}
\end{quote}

\paragraph{Example paraphrases.}
Table~\ref{tab:libero_paraphrase_examples} shows representative LIBERO-Spatial language contrast-set paraphrases. Each paraphrase preserves the target object, action, and goal location while changing the surface form of the instruction.

\begin{table}[htbp]
\centering
\small
\caption{Representative language contrast-set paraphrases for LIBERO-Spatial tasks.}
\label{tab:libero_paraphrase_examples}
\resizebox{\linewidth}{!}{%
\begin{tabular}{p{0.32\linewidth}p{0.32\linewidth}p{0.32\linewidth}}
\toprule
Paraphrase 1 & Paraphrase 2 & Paraphrase 3 \\
\midrule
Lift the black bowl that is placed between the plate and the ramekin, and set it on the plate.
& Pick up the black bowl located between the plate and the ramekin, then move it onto the plate.
& Grab the black bowl sitting between the ramekin and the plate and place it onto the plate. \\

\hdashline

Pick up the black bowl from the center of the table and place it on the plate.
& Lift the black bowl located at the table center and set it onto the plate.
& Grab the black bowl from the middle of the table and put it on the plate. \\

\hdashline
Lift the black bowl from the cookie box and place it on the plate.
& Pick up the black bowl sitting on top of the cookie box and move it to the plate.
& Grab the black bowl resting on the cookie box and set it on the plate. \\

\hdashline
Pick up the black bowl from the top drawer of the wooden cabinet and place it on the plate.
& Lift the black bowl inside the upper drawer of the wooden cabinet and move it onto the plate.
& Retrieve the black bowl from the wooden cabinet top drawer and put it on the plate. \\
\hdashline
Lift the black bowl on the ramekin and place it on the plate.
& Pick up the black bowl resting on the ramekin and move it onto the plate.
& Grab the black bowl sitting on top of the ramekin and set it onto the plate. \\
\bottomrule
\end{tabular}%
}
\end{table}

\paragraph{Real-world perturbations.}
In the DROID/Franka setup, visual contrast sets include distractor objects, cluttered backgrounds, novel objects, and altered object configurations. Language contrast sets include paraphrased instructions and negated or distractor phrasing when applicable. Joint visual-language contrast sets combine these visual and language changes, inducing multimodal deployment shifts. Table~\ref{tab:droid_task_examples} shows representative real-world source and contrast-set conditions.

\begin{table}[htbp]
\centering
\small
\caption{Examples of real-world DROID source and contrast-set conditions.}
\label{tab:droid_task_examples}
\resizebox{\linewidth}{!}{%
\begin{tabular}{l|l|l}
\toprule
Source Task & Language Perturbation & Visual Perturbation \\
\midrule
\multirow{5}{*}{pick up the cube and place it in the bowl}
& grab the cube and drop it inside the bowl & Distractor: Plushie \\
& lift the cube and put it into the bowl & Distractor: Marker \\
& & Distractor: Marker and Plushie \\
& & Black background texture \\
& & White background texture \\
\midrule
\multirow{3}{*}{pick up the marker and place it in the bowl}
& grab the marker and place it inside the bowl & Distractor: Additional object near target \\
& pick up the marker and set it in the bowl & Background texture variation \\
& take the marker near the cup and place it in the bowl & Additional object near cup \\
\bottomrule
\end{tabular}%
}
\end{table}


\section{Experiment Setup Details}
\label{app:experiment_setup}

This section provides additional setup details for the simulation and real-world experiments, including perturbation procedures, dataset filtering, and task conditions.

\subsection{LIBERO Simulation Setup}
\label{app:libero_setup}

In simulation, source-distribution rollouts are collected from LIBERO-Spatial. Evaluation is performed on LIBERO-Plus, which contains deployment-time shifts relative to the source distribution. Contrast-set rollouts are generated by perturbing LIBERO-Spatial source conditions through visual distractors or language paraphrases, then re-executing the frozen VLA policy.

\subsection{DROID Real-World Setup}
\label{app:droid_setup}

In the real-world setup, $\pi_0$ and $\pi_0$-FAST are evaluated on spatial pick-and-place tasks using a Franka/DROID setup. The source-distribution rollouts use fixed task families, while contrast-set and evaluation rollouts introduce visual, language, and joint visual-language shifts.

\subsection{Dynamic Time Warping Filtering}
\label{app:dtw_filtering}
\label{dtw}

For probe training, candidate contrast-set rollouts are filtered using an active rejection procedure based on Dynamic Time Warping (DTW). The goal of this filtering step is to reduce near-duplicate trajectories in the contrast-set training pool and increase diversity among selected perturbation rollouts. Given a candidate rollout, we compute its DTW distance to already-selected contrast-set rollouts from the same task and outcome class. If the candidate is too similar to an existing selected rollout under a fixed DTW distance threshold, it is rejected; otherwise, it is added to the training pool.

\paragraph{Trajectory representation.}
Each rollout is represented by the time series of 3D robot end-effector positions at every control step. Before DTW, each sequence is translation-normalized by subtracting its initial end-effector position.

\paragraph{DTW distance and threshold.}
We compute FastDTW with Euclidean distance between 3D end-effector waypoints and divide the resulting distance by the optimal warping path length. A candidate is accepted only if its minimum normalized DTW distance to accepted rollouts in the same task-and-outcome pool is at least
\[
\tau_{\mathrm{DTW}} = 0.02.
\]
We use $\tau_{\mathrm{DTW}} = 0.02$ during LIBERO contrast-set rollout collection for both OpenVLA and $\pi_0$.

\begin{algorithm}[t]
\caption{DTW-based active rejection for contrast-set rollout collection}
\label{alg:dtw_active_rejection}
\begin{algorithmic}[1]
\Require Task $t$, outcome budgets $K_{\mathrm{succ}}$ and $K_{\mathrm{fail}}$, threshold $\tau_{\mathrm{DTW}} = 0.02$
\State Initialize accepted pools $\mathcal{A}_{t,\mathrm{succ}} \gets \emptyset$, $\mathcal{A}_{t,\mathrm{fail}} \gets \emptyset$
\While{$|\mathcal{A}_{t,\mathrm{succ}}| < K_{\mathrm{succ}}$ \textbf{or} $|\mathcal{A}_{t,\mathrm{fail}}| < K_{\mathrm{fail}}$}
    \State Execute policy under the perturbed contrast-set condition to obtain rollout $\tau$ and label $y \in \{\mathrm{succ}, \mathrm{fail}\}$
    \State Form end-effector trajectory $\tilde{\tau} \gets \mathrm{NormalizeEE}(\tau)$ by subtracting the initial position
    \If{$|\mathcal{A}_{t,y}| < K_y$}
        \If{$\mathcal{A}_{t,y} = \emptyset$ \textbf{or} $\min_{\tilde{\tau}_j \in \mathcal{A}_{t,y}} d_{\mathrm{DTW}}(\tilde{\tau}, \tilde{\tau}_j) \ge \tau_{\mathrm{DTW}}$}
            \State Accept $\tau$ and set $\mathcal{A}_{t,y} \gets \mathcal{A}_{t,y} \cup \{\tilde{\tau}\}$
        \Else
            \State Reject $\tau$
        \EndIf
    \Else
        \State Reject $\tau$
    \EndIf
\EndWhile
\end{algorithmic}
\end{algorithm}

\paragraph{Active-rejection procedure.}
Algorithm~\ref{alg:dtw_active_rejection} summarizes the active rejection procedure. Success and failure pools are maintained separately for each task; rejected rollouts are not saved to $\DcsTrain$. This produces the contrast-set training pool
\[
\DcsTrain = \DcsVisTrain \cup \DcsLangTrain \cup \DcsVLTrain,
\]
and the augmented probe-training set
\[
\DaugTrain = \DsrcTrain \cup \DcsTrain.
\]
For conformal calibration, the contrast-set calibration pool is
\[
\DcsCal = \DcsVisCal \cup \DcsLangCal \cup \DcsVLCal,
\]
and the augmented calibration set is
\[
\DaugCal = \DsrcCal \cup \DcsCal.
\]
Calibration subsets are sampled with controlled success and failure counts where possible so that threshold estimation is not dominated by a single outcome class.




\section{Statistical Testing Details}
\label{app:stats}

We test whether contrast-set-aware configurations improve failure detection relative to the source-distribution baseline using seed-paired $t$-tests on full-curve functional CP ROC-AUC. Each comparison reports the mean difference relative to $\Cone$, with positive values indicating improvement over the baseline. We apply Bonferroni correction over $m=12$ tests. Table \ref{tab:paired-tests-roc-alpha-combined-sim-real} enumerates the p values.

\begin{table}[t]
  \centering
  \caption{Seed-paired $t$-tests for full-curve functional CP ROC-AUC over $\alpha \in \{0.1, \ldots, 0.9\}$. Each cell reports mean difference (other $-$ $\Cone$) and raw $p$-value. \textsuperscript{*} marks Bonferroni significance over $m=12$ tests.}
  \label{tab:paired-tests-roc-alpha-combined-sim-real}
  \small
  \setlength{\tabcolsep}{4pt}
  \renewcommand{\arraystretch}{1.08}
  \resizebox{\linewidth}{!}{%
  \begin{tabular}{l|cccc}
    \toprule
    Comparison & Sim-$\pi_0$ & Sim-OpenVLA & Real-$\pi_0$ & Real-$\pi_0$-FAST \\
    \midrule
    $\Ctwo$ vs.\ $\Cone$
    & \textbf{+0.1011}\textsuperscript{*} ($p<10^{-10}$)
    & +0.0647 ($p=0.4093$)
    & +0.0023 ($p=0.8753$)
    & \textbf{+0.1120}\textsuperscript{*} ($p<10^{-5}$) \\
    $\Cthree$ vs.\ $\Cone$
    & \textbf{+0.1115}\textsuperscript{*} ($p<10^{-10}$)
    & \textbf{+0.0773}\textsuperscript{*} ($p=0.0008$)
    & \textbf{+0.0789}\textsuperscript{*} ($p<10^{-10}$)
    & \textbf{+0.1077}\textsuperscript{*} ($p<10^{-7}$) \\
    \Cfour{} vs.\ \Cone
    & \textbf{+0.1202}\textsuperscript{*} ($p<10^{-10}$)
    & +0.2023 ($p=0.0256$)
    & \textbf{+0.1181}\textsuperscript{*} ($p<10^{-8}$)
    & \textbf{+0.1162}\textsuperscript{*} ($p<10^{-7}$) \\
    \bottomrule
  \end{tabular}%
  }
\end{table}


\section{Additional Alpha Sweep Results}
\label{app:alpha_sweeps}

This section reports the full conformal-significance sweep used to support the $\alpha$-marginalized results in the main paper. For each setting, we report both F1 and ROC-AUC. F1 measures calibrated rollout-level failure detection quality at each operating point, while ROC-AUC measures separability between successful and failed rollouts induced by the conformal threshold sweep. 

\subsection{Simulation Alpha Sweep Results}
\label{app:sim_alpha}

Tables~\ref{tab:f1-by-alpha-pizero-openvla} and~\ref{tab:roc-auc-by-alpha-pizero-openvla} report simulation performance for $\pi_0$ and OpenVLA on LIBERO. The F1 table shows the practical detection quality at each risk tolerance, while the ROC-AUC table shows how the conformal threshold sweep separates successful and failed rollouts. Relative gains are computed with respect to $\Cone$ at the same $\alpha$.

\begin{table*}[!htbp]
  \centering
  \caption{
  F1 vs.\ $\alpha$ for $\pi_0$ and OpenVLA in the LIBERO simulation suite, averaged over 30 seeds. 
  The Relative gain $\Delta C_k = \mathrm{F1}(C_k) - \mathrm{F1}(\Cone)$.
  C1=$\Cone$, C2=$\Ctwo$, C3=$\Cthree$, and C4=$\Cfour$.
  }
  \label{tab:f1-by-alpha-pizero-openvla}
  \scriptsize
  \setlength{\tabcolsep}{2.5pt}
  \renewcommand{\arraystretch}{0.95}
  \resizebox{\textwidth}{!}{%
  \begin{tabular}{c|cccc|ccc|cccc|ccc}
    \toprule
    & \multicolumn{7}{c|}{\textbf{$\pi_0$}} 
    & \multicolumn{7}{c}{\textbf{OpenVLA}} \\
    \cmidrule(lr){2-8}\cmidrule(lr){9-15}
    $\alpha$
    & C1 & C2 & C3 & C4
    & $\Delta$C2 & $\Delta$C3 & $\Delta$C4
    & C1 & C2 & C3 & C4
    & $\Delta$C2 & $\Delta$C3 & $\Delta$C4 \\
    \midrule
    0.02 & 0.880 & 0.861 & 0.836 & \textbf{0.944} & -0.019 & -0.044 & \textbf{+0.064} & 0.841 & \textbf{0.930} & 0.731 & 0.767 & \textbf{+0.089} & -0.110 & -0.074 \\
    0.05 & 0.845 & 0.890 & 0.841 & \textbf{0.914} & +0.045 & -0.004 & \textbf{+0.069} & 0.856 & \textbf{0.929} & 0.809 & 0.863 & \textbf{+0.072} & -0.048 & +0.007 \\
    0.10 & 0.807 & 0.883 & 0.880 & \textbf{0.895} & +0.077 & +0.074 & \textbf{+0.088} & 0.877 & 0.922 & 0.878 & \textbf{0.942} & +0.045 & +0.001 & \textbf{+0.065} \\
    0.15 & 0.786 & 0.859 & 0.861 & \textbf{0.871} & +0.074 & +0.075 & \textbf{+0.086} & 0.889 & 0.918 & 0.915 & \textbf{0.936} & +0.028 & +0.026 & \textbf{+0.047} \\
    0.20 & 0.769 & 0.837 & 0.828 & \textbf{0.846} & +0.069 & +0.060 & \textbf{+0.077} & 0.892 & 0.907 & 0.908 & \textbf{0.919} & +0.015 & +0.016 & \textbf{+0.027} \\
    0.25 & 0.755 & 0.809 & 0.796 & \textbf{0.820} & +0.054 & +0.041 & \textbf{+0.065} & 0.885 & 0.898 & 0.896 & \textbf{0.906} & +0.013 & +0.012 & \textbf{+0.021} \\
    0.30 & 0.736 & 0.790 & 0.772 & \textbf{0.797} & +0.053 & +0.036 & \textbf{+0.060} & 0.876 & 0.887 & 0.887 & \textbf{0.892} & +0.011 & +0.011 & \textbf{+0.016} \\
    0.35 & 0.729 & 0.774 & 0.752 & \textbf{0.778} & +0.045 & +0.023 & \textbf{+0.049} & 0.870 & 0.877 & 0.880 & \textbf{0.881} & +0.007 & +0.010 & \textbf{+0.011} \\
    0.40 & 0.724 & 0.762 & 0.737 & \textbf{0.760} & +0.037 & +0.013 & \textbf{+0.036} & 0.866 & 0.871 & 0.871 & \textbf{0.873} & +0.005 & +0.005 & \textbf{+0.007} \\
    0.45 & 0.715 & 0.752 & 0.722 & \textbf{0.747} & +0.037 & +0.008 & \textbf{+0.032} & 0.861 & 0.864 & 0.865 & \textbf{0.865} & +0.004 & \textbf{+0.005} & +0.004 \\
    0.50 & 0.704 & 0.741 & 0.715 & \textbf{0.733} & +0.037 & +0.011 & \textbf{+0.029} & 0.856 & 0.859 & 0.859 & \textbf{0.859} & +0.003 & \textbf{+0.003} & +0.002 \\
    0.60 & 0.689 & 0.718 & 0.697 & \textbf{0.712} & +0.029 & +0.008 & \textbf{+0.023} & 0.847 & 0.851 & 0.849 & \textbf{0.851} & +0.003 & +0.002 & \textbf{+0.004} \\
    0.70 & 0.676 & 0.696 & 0.683 & \textbf{0.695} & +0.020 & +0.006 & \textbf{+0.019} & 0.842 & 0.845 & 0.841 & \textbf{0.845} & \textbf{+0.004} & -0.000 & +0.003 \\
    0.80 & 0.662 & 0.678 & 0.670 & \textbf{0.679} & +0.016 & +0.008 & \textbf{+0.017} & 0.838 & 0.840 & 0.837 & \textbf{0.839} & \textbf{+0.002} & -0.001 & +0.001 \\
    0.90 & 0.652 & 0.660 & 0.653 & \textbf{0.658} & +0.008 & +0.001 & \textbf{+0.006} & 0.835 & \textbf{0.836} & 0.835 & 0.835 & \textbf{+0.001} & -0.000 & +0.000 \\
    \bottomrule
  \end{tabular}%
  }
\end{table*}

\begin{table*}[!htbp]
  \centering
  \caption{
  Cumulative ROC-AUC vs.\ $\alpha$ for $\pi_0$ and OpenVLA in the LIBERO simulation suite, averaged over 30 seeds. The Relative Gain
  $\Delta C_k = \mathrm{ROC}(C_k) - \mathrm{ROC}(\Cone)$.
  C1=$\Cone$, C2=$\Ctwo$, C3=$\Cthree$, and C4=$\Cfour$.
  }
  \label{tab:roc-auc-by-alpha-pizero-openvla}
  \scriptsize
  \setlength{\tabcolsep}{2.5pt}
  \renewcommand{\arraystretch}{0.95}
  \resizebox{\textwidth}{!}{%
  \begin{tabular}{c|cccc|ccc|cccc|ccc}
    \toprule
    & \multicolumn{7}{c|}{\textbf{$\pi_0$}} 
    & \multicolumn{7}{c}{\textbf{OpenVLA}} \\
    \cmidrule(lr){2-8}\cmidrule(lr){9-15}
    $\alpha$
    & C1 & C2 & C3 & C4
    & $\Delta$C2 & $\Delta$C3 & $\Delta$C4
    & C1 & C2 & C3 & C4
    & $\Delta$C2 & $\Delta$C3 & $\Delta$C4 \\
    \midrule
    0.02 & \textbf{0.000} & \textbf{0.000} & \textbf{0.000} & \textbf{0.000} & +0.000 & +0.000 & +0.000 & \textbf{0.000} & \textbf{0.000} & \textbf{0.000} & \textbf{0.000} & +0.000 & +0.000 & +0.000 \\
    0.05 & \textbf{0.053} & 0.039 & 0.038 & 0.039 & -0.014 & -0.015 & -0.014 & 0.036 & \textbf{0.046} & 0.010 & 0.019 & +0.010 & -0.025 & -0.016 \\
    0.10 & \textbf{0.116} & 0.085 & 0.065 & 0.067 & -0.031 & -0.051 & -0.050 & 0.076 & \textbf{0.154} & 0.042 & 0.083 & +0.077 & -0.034 & +0.007 \\
    0.15 & \textbf{0.155} & 0.123 & 0.090 & 0.099 & -0.033 & -0.065 & -0.056 & 0.137 & \textbf{0.242} & 0.094 & 0.178 & +0.106 & -0.043 & +0.041 \\
    0.20 & \textbf{0.190} & 0.156 & 0.137 & 0.138 & -0.034 & -0.053 & -0.052 & 0.201 & \textbf{0.322} & 0.160 & 0.313 & +0.121 & -0.041 & +0.112 \\
    0.25 & \textbf{0.221} & 0.203 & 0.193 & 0.180 & -0.017 & -0.028 & -0.041 & 0.267 & 0.387 & 0.243 & \textbf{0.393} & +0.120 & -0.024 & +0.126 \\
    0.30 & \textbf{0.254} & 0.239 & 0.234 & 0.219 & -0.015 & -0.020 & -0.035 & 0.341 & 0.463 & 0.313 & \textbf{0.483} & +0.123 & -0.027 & +0.142 \\
    0.35 & \textbf{0.274} & 0.268 & 0.273 & 0.252 & -0.006 & -0.001 & -0.022 & 0.380 & 0.526 & 0.365 & \textbf{0.551} & +0.146 & -0.015 & +0.171 \\
    0.40 & 0.288 & 0.292 & \textbf{0.307} & 0.286 & +0.004 & +0.019 & -0.002 & 0.412 & 0.567 & 0.424 & \textbf{0.604} & +0.155 & +0.012 & +0.192 \\
    0.45 & 0.304 & 0.311 & \textbf{0.336} & 0.312 & +0.008 & +0.032 & +0.009 & 0.453 & 0.609 & 0.460 & \textbf{0.666} & +0.156 & +0.008 & +0.213 \\
    0.50 & 0.322 & 0.331 & \textbf{0.352} & 0.339 & +0.009 & +0.030 & +0.017 & 0.487 & 0.647 & 0.503 & \textbf{0.709} & +0.161 & +0.016 & +0.223 \\
    0.60 & 0.355 & 0.377 & \textbf{0.388} & 0.384 & +0.022 & +0.034 & +0.029 & 0.542 & 0.707 & 0.586 & \textbf{0.761} & +0.165 & +0.045 & +0.219 \\
    0.70 & 0.383 & \textbf{0.424} & 0.422 & 0.422 & +0.041 & +0.039 & +0.039 & 0.595 & 0.742 & 0.644 & \textbf{0.812} & +0.147 & +0.049 & +0.216 \\
    0.80 & 0.415 & \textbf{0.467} & 0.453 & 0.461 & +0.052 & +0.038 & +0.046 & 0.631 & 0.781 & 0.693 & \textbf{0.853} & +0.150 & +0.062 & +0.222 \\
    0.90 & 0.440 & \textbf{0.512} & 0.494 & 0.511 & +0.071 & +0.054 & +0.071 & 0.670 & 0.807 & 0.725 & \textbf{0.883} & +0.137 & +0.055 & +0.213 \\
    \bottomrule
  \end{tabular}%
  }
\end{table*}

\subsection{Real-World DROID Alpha Sweep Results}
\label{app:droid_alpha}
\label{app:droid_alpha_roc}

Tables~\ref{tab:f1-by-alpha-pi0-pi0fast-droid} and~\ref{tab:roc-auc-by-alpha-pi0-pi0fast-droid} report the real-world DROID alpha sweeps for $\pi_0$ and $\pi_0$-FAST. The F1 table summarizes calibrated failure-detection quality at each deployment risk tolerance, while the ROC-AUC table summarizes the corresponding success/failure separability induced by the threshold sweep.

\begin{table*}[!htbp]
  \centering
  \caption{
  F1 vs.\ $\alpha$ for $\pi_0$ and $\pi_0$-FAST on DROID real-world rollouts, averaged over 30 seeds. 
  $\Delta C_k = \mathrm{F1}(C_k) - \mathrm{F1}(\Cone)$.
  C1=$\Cone$, C2=$\Ctwo$, C3=$\Cthree$, and C4=$\Cfour$.
  }
  \label{tab:f1-by-alpha-pi0-pi0fast-droid}
  \scriptsize
  \setlength{\tabcolsep}{2.5pt}
  \renewcommand{\arraystretch}{0.95}
  \resizebox{\textwidth}{!}{%
  \begin{tabular}{c|cccc|ccc|cccc|ccc}
    \toprule
    & \multicolumn{7}{c|}{\textbf{$\pi_0$}} 
    & \multicolumn{7}{c}{\textbf{$\pi_0$-FAST}} \\
    \cmidrule(lr){2-8}\cmidrule(lr){9-15}
    $\alpha$
    & C1 & C2 & C3 & C4
    & $\Delta$C2 & $\Delta$C3 & $\Delta$C4
    & C1 & C2 & C3 & C4
    & $\Delta$C2 & $\Delta$C3 & $\Delta$C4 \\
    \midrule
    0.02 & 0.148 & \textbf{0.388} & 0.108 & 0.203 & \textbf{+0.240} & -0.040 & +0.055 & 0.567 & \textbf{0.703} & 0.543 & 0.615 & \textbf{+0.137} & -0.024 & +0.049 \\
    0.05 & 0.153 & \textbf{0.413} & 0.128 & 0.244 & \textbf{+0.260} & -0.025 & +0.091 & 0.594 & \textbf{0.731} & 0.581 & 0.641 & \textbf{+0.136} & -0.014 & +0.046 \\
    0.10 & 0.169 & \textbf{0.453} & 0.178 & 0.303 & \textbf{+0.284} & +0.008 & +0.134 & 0.627 & \textbf{0.770} & 0.621 & 0.686 & \textbf{+0.143} & -0.006 & +0.059 \\
    0.15 & 0.206 & \textbf{0.483} & 0.242 & 0.338 & \textbf{+0.277} & +0.036 & +0.133 & 0.662 & \textbf{0.791} & 0.656 & 0.736 & \textbf{+0.129} & -0.006 & +0.074 \\
    0.20 & 0.256 & \textbf{0.504} & 0.287 & 0.426 & \textbf{+0.248} & +0.031 & +0.170 & 0.693 & 0.806 & 0.707 & \textbf{0.807} & +0.113 & +0.014 & \textbf{+0.113} \\
    0.25 & 0.300 & \textbf{0.538} & 0.329 & 0.467 & \textbf{+0.239} & +0.029 & +0.167 & 0.710 & 0.813 & 0.732 & \textbf{0.822} & +0.103 & +0.022 & \textbf{+0.111} \\
    0.30 & 0.331 & \textbf{0.556} & 0.357 & 0.485 & \textbf{+0.225} & +0.026 & +0.154 & 0.724 & 0.827 & 0.768 & \textbf{0.834} & +0.103 & +0.045 & \textbf{+0.110} \\
    0.35 & 0.361 & \textbf{0.576} & 0.393 & 0.500 & \textbf{+0.215} & +0.031 & +0.139 & 0.757 & \textbf{0.838} & 0.778 & 0.837 & \textbf{+0.081} & +0.021 & +0.080 \\
    0.40 & 0.372 & \textbf{0.583} & 0.421 & 0.525 & \textbf{+0.211} & +0.049 & +0.153 & 0.773 & \textbf{0.837} & 0.787 & 0.835 & \textbf{+0.064} & +0.014 & +0.062 \\
    0.45 & 0.412 & \textbf{0.595} & 0.442 & 0.550 & \textbf{+0.183} & +0.030 & +0.138 & 0.777 & \textbf{0.838} & 0.780 & 0.836 & \textbf{+0.061} & +0.002 & +0.059 \\
    0.50 & 0.431 & \textbf{0.608} & 0.465 & 0.581 & \textbf{+0.177} & +0.034 & +0.150 & 0.779 & 0.839 & 0.777 & \textbf{0.841} & +0.060 & -0.002 & \textbf{+0.062} \\
    0.60 & 0.459 & \textbf{0.617} & 0.522 & 0.604 & \textbf{+0.158} & +0.062 & +0.145 & 0.785 & 0.825 & 0.784 & \textbf{0.830} & +0.039 & -0.002 & \textbf{+0.044} \\
    0.70 & 0.514 & \textbf{0.632} & 0.567 & 0.632 & \textbf{+0.118} & +0.052 & +0.117 & 0.783 & 0.807 & 0.791 & \textbf{0.815} & +0.025 & +0.008 & \textbf{+0.033} \\
    0.80 & 0.553 & 0.646 & 0.602 & \textbf{0.650} & +0.093 & +0.049 & \textbf{+0.096} & 0.785 & 0.799 & 0.784 & \textbf{0.805} & +0.015 & -0.001 & \textbf{+0.020} \\
    0.90 & 0.591 & 0.654 & 0.635 & \textbf{0.665} & +0.063 & +0.045 & \textbf{+0.074} & 0.774 & 0.790 & 0.769 & \textbf{0.796} & +0.016 & -0.005 & \textbf{+0.022} \\
    \bottomrule
  \end{tabular}%
  }
\end{table*}

\begin{table*}[!htbp]
  \centering
  \caption{
  Cumulative ROC-AUC vs.\ $\alpha$ for $\pi_0$ and $\pi_0$-FAST on DROID real-world rollouts, averaged over 30 seeds. 
  $\Delta C_k = \mathrm{ROC}(C_k) - \mathrm{ROC}(\Cone)$.
  C1=$\Cone$, C2=$\Ctwo$, C3=$\Cthree$, and C4=$\Cfour$.
  }
  \label{tab:roc-auc-by-alpha-pi0-pi0fast-droid}
  \scriptsize
  \setlength{\tabcolsep}{2.5pt}
  \renewcommand{\arraystretch}{0.95}
  \resizebox{\textwidth}{!}{%
  \begin{tabular}{c|cccc|ccc|cccc|ccc}
    \toprule
    & \multicolumn{7}{c|}{\textbf{$\pi_0$}} 
    & \multicolumn{7}{c}{\textbf{$\pi_0$-FAST}} \\
    \cmidrule(lr){2-8}\cmidrule(lr){9-15}
    $\alpha$
    & C1 & C2 & C3 & C4
    & $\Delta$C2 & $\Delta$C3 & $\Delta$C4
    & C1 & C2 & C3 & C4
    & $\Delta$C2 & $\Delta$C3 & $\Delta$C4 \\
    \midrule
    0.02 & \textbf{0.000} & \textbf{0.000} & \textbf{0.000} & \textbf{0.000} & +0.000 & +0.000 & +0.000 & \textbf{0.000} & \textbf{0.000} & \textbf{0.000} & \textbf{0.000} & +0.000 & +0.000 & +0.000 \\
    0.05 & 0.001 & \textbf{0.023} & 0.001 & 0.003 & +0.022 & +0.000 & +0.003 & 0.007 & \textbf{0.014} & 0.007 & 0.007 & +0.007 & +0.000 & +0.000 \\
    0.10 & 0.002 & \textbf{0.058} & 0.004 & 0.014 & +0.055 & +0.002 & +0.012 & 0.021 & \textbf{0.046} & 0.028 & 0.036 & +0.025 & +0.007 & +0.015 \\
    0.15 & 0.008 & \textbf{0.089} & 0.015 & 0.031 & +0.081 & +0.007 & +0.023 & 0.043 & \textbf{0.069} & 0.052 & 0.054 & +0.026 & +0.008 & +0.011 \\
    0.20 & 0.015 & \textbf{0.114} & 0.024 & 0.066 & +0.099 & +0.009 & +0.051 & 0.071 & \textbf{0.092} & 0.073 & 0.080 & +0.021 & +0.002 & +0.009 \\
    0.25 & 0.027 & \textbf{0.147} & 0.041 & 0.096 & +0.119 & +0.013 & +0.068 & 0.101 & \textbf{0.119} & 0.087 & 0.104 & +0.018 & -0.014 & +0.003 \\
    0.30 & 0.045 & \textbf{0.170} & 0.059 & 0.114 & +0.125 & +0.015 & +0.069 & 0.136 & \textbf{0.138} & 0.115 & 0.122 & +0.001 & -0.021 & -0.015 \\
    0.35 & 0.062 & \textbf{0.197} & 0.084 & 0.141 & +0.136 & +0.022 & +0.079 & 0.175 & \textbf{0.187} & 0.147 & 0.157 & +0.013 & -0.028 & -0.017 \\
    0.40 & 0.072 & \textbf{0.209} & 0.107 & 0.174 & +0.137 & +0.034 & +0.101 & 0.204 & \textbf{0.225} & 0.177 & 0.222 & +0.021 & -0.027 & +0.018 \\
    0.45 & 0.092 & \textbf{0.235} & 0.124 & 0.199 & +0.143 & +0.032 & +0.108 & 0.249 & 0.272 & 0.244 & \textbf{0.284} & +0.023 & -0.005 & +0.035 \\
    0.50 & 0.103 & \textbf{0.249} & 0.137 & 0.230 & +0.146 & +0.035 & +0.127 & 0.300 & 0.356 & 0.319 & \textbf{0.364} & +0.057 & +0.019 & +0.065 \\
    0.60 & 0.132 & 0.262 & 0.190 & \textbf{0.273} & +0.130 & +0.058 & +0.141 & 0.372 & 0.487 & 0.407 & \textbf{0.503} & +0.115 & +0.035 & +0.131 \\
    0.70 & 0.167 & 0.290 & 0.245 & \textbf{0.325} & +0.124 & +0.078 & +0.158 & 0.422 & \textbf{0.607} & 0.463 & 0.605 & +0.185 & +0.041 & +0.183 \\
    0.80 & 0.225 & 0.308 & 0.288 & \textbf{0.368} & +0.083 & +0.063 & +0.143 & 0.470 & 0.655 & 0.544 & \textbf{0.659} & +0.185 & +0.074 & +0.189 \\
    0.90 & 0.265 & 0.322 & 0.345 & \textbf{0.394} & +0.057 & +0.081 & +0.129 & 0.572 & \textbf{0.709} & 0.682 & 0.701 & +0.136 & +0.110 & +0.129 \\
    \bottomrule
  \end{tabular}%
  }
\end{table*}


\section{Additional Ablations}
\label{app:ablations}

\subsection{Training Exposure vs. Calibration Alignment}
\label{app:training_vs_calibration}

This ablation compares $\Ctwo$ and $\Cthree$ to isolate the relative effects of contrast-set exposure during probe training and contrast-set alignment during conformal calibration. $\Ctwo$ changes the learned failure-score function by training the probe on source and contrast-set rollouts, while $\Cthree$ keeps the probe trained on source rollouts but estimates conformal thresholds using augmented calibration trajectories. The results indicate that both mechanisms can improve robustness, but the dominant mechanism varies across policy backbones and deployment settings.

\subsection{Additional Modality Analysis}
\label{app:modality}

The main paper reports F1 trends for visual-only, language-only, and joint visual-language contrast-set ablations. Table~\ref{tab:modality_roc_auc_appendix} summarizes the corresponding $\alpha$-marginal cumulative ROC-AUC.

\begin{table}[htbp]
  \centering
  \small
  \caption{$\alpha$-marginal cumulative ROC-AUC for contrast-set modality ablations.}
  \label{tab:modality_roc_auc_appendix}
  \begin{tabular}{lcc}
    \toprule
    Contrast sets & OpenVLA ROC-AUC $\uparrow$ & $\pi_0$ ROC-AUC $\uparrow$ \\
    \midrule
    Visual only & 0.484 & 0.469 \\
    Language only & 0.484 & 0.441 \\
    Visual + Language ($\Cfour$) & \textbf{0.561} & 0.282 \\
    \bottomrule
  \end{tabular}
\end{table}


\subsection{Additional Sim-to-Real Analysis}
\label{app:sim2real}

The main paper evaluates a simulation-trained probe calibrated on real-world contrast-set trajectories. Table~\ref{tab:sim2real_appendix} compares $\alpha$-marginal F1 and cumulative ROC-AUC under three probe/calibration sources. The sim-only row reports in-simulation $\Cfour{}$ performance; the transfer row uses a simulation-trained $\pi_0$ probe with real contrast-set calibration and evaluates on real DROID rollouts.

\begin{table}[htbp]
  \centering
  \small
  \caption{$\alpha$-marginal failure detection on real DROID rollouts under different probe/calibration sources.}
  \label{tab:sim2real_appendix}
  \begin{tabular}{lcc}
    \toprule
    Configuration & F1 $\uparrow$ & ROC-AUC $\uparrow$ \\
    \midrule
    Real-only probe and calibration & 0.478 & 0.228 \\
    Sim-trained probe + real CS calibration & \textbf{0.523} & \textbf{0.410} \\
    \bottomrule
  \end{tabular}
\end{table}

\subsection{Low- and High-Alpha Regimes}
\label{app:low_high_alpha}

The full $\alpha$ sweeps in Appendix~\ref{app:alpha_sweeps} summarize performance across deployment risk tolerances. 
Here, we isolate low- and high-$\alpha$ regimes to show how methods behave under permissive versus conservative intervention settings. 
In our functional CP setup, higher $\alpha$ values correspond to lower risk thresholds and therefore more conservative intervention, while lower $\alpha$ values are more permissive and reduce unnecessary alarms.

\subsubsection{Simulation Results}
\label{app:low_high_alpha_sim}

\begin{figure}[t]
    \centering
    \includegraphics[width=\linewidth]{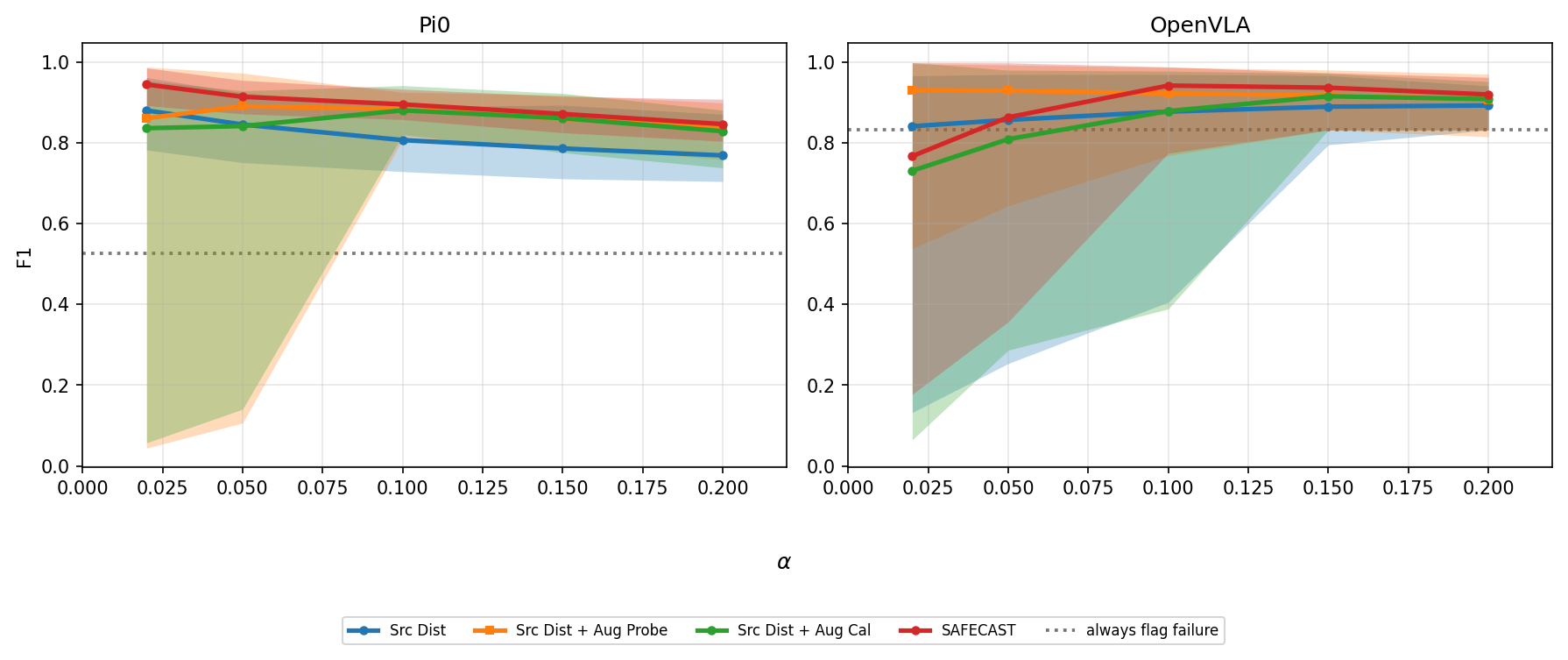}
    \caption{
    Simulation failure detection in the low-$\alpha$ regime, where thresholds are more permissive and fewer rollouts are flagged.
    }
    \label{fig:sim_low_alpha}
\end{figure}

\begin{figure}[t]
    \centering
    \includegraphics[width=\linewidth]{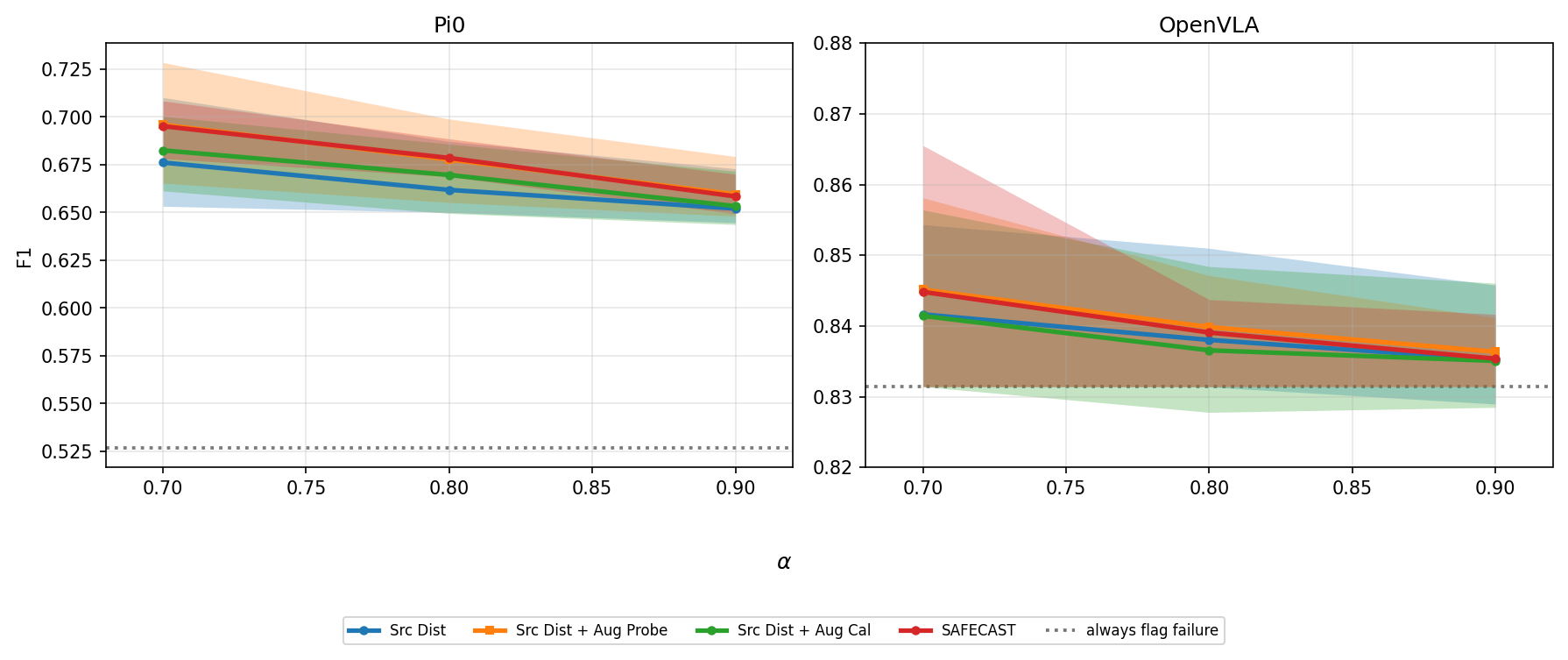}
    \caption{
    Simulation failure detection in the high-$\alpha$ regime, where thresholds are more conservative and failures are flagged more aggressively.
    }
    \label{fig:sim_high_alpha}
\end{figure}

Figures~\ref{fig:sim_low_alpha} and~\ref{fig:sim_high_alpha} show simulation performance in the low- and high-$\alpha$ regimes. 
The high-$\alpha$ regime is most relevant for safety-critical deployment because it emphasizes conservative failure detection under shifted conditions. 
The low-$\alpha$ regime provides a complementary view of whether methods remain stable when fewer interventions are preferred.

\subsubsection{Real-World Results}
\label{app:low_high_alpha_real}

\begin{figure}[t]
    \centering
    \includegraphics[width=\linewidth]{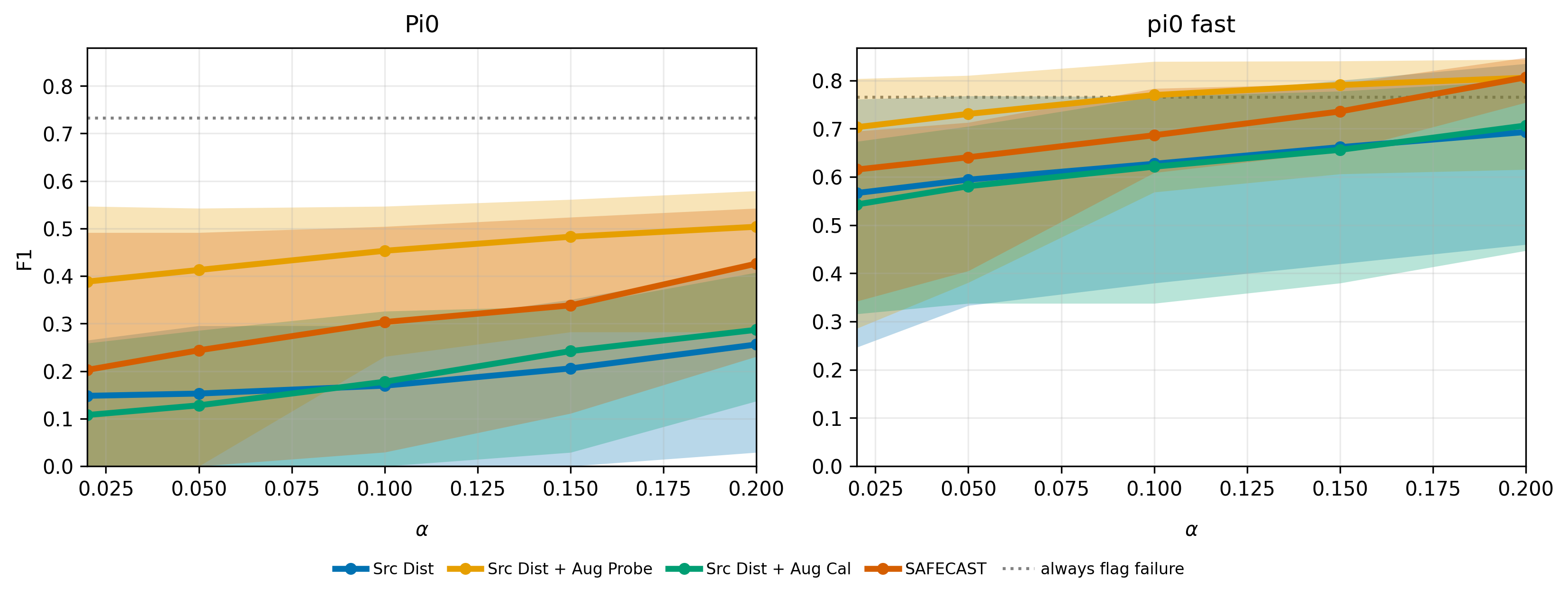}
    \caption{
    Real-world failure detection in the low-$\alpha$ regime, emphasizing permissive operation with fewer interventions.
    }
    \label{fig:real_low_alpha}
\end{figure}

\begin{figure}[t]
    \centering
    \includegraphics[width=\linewidth]{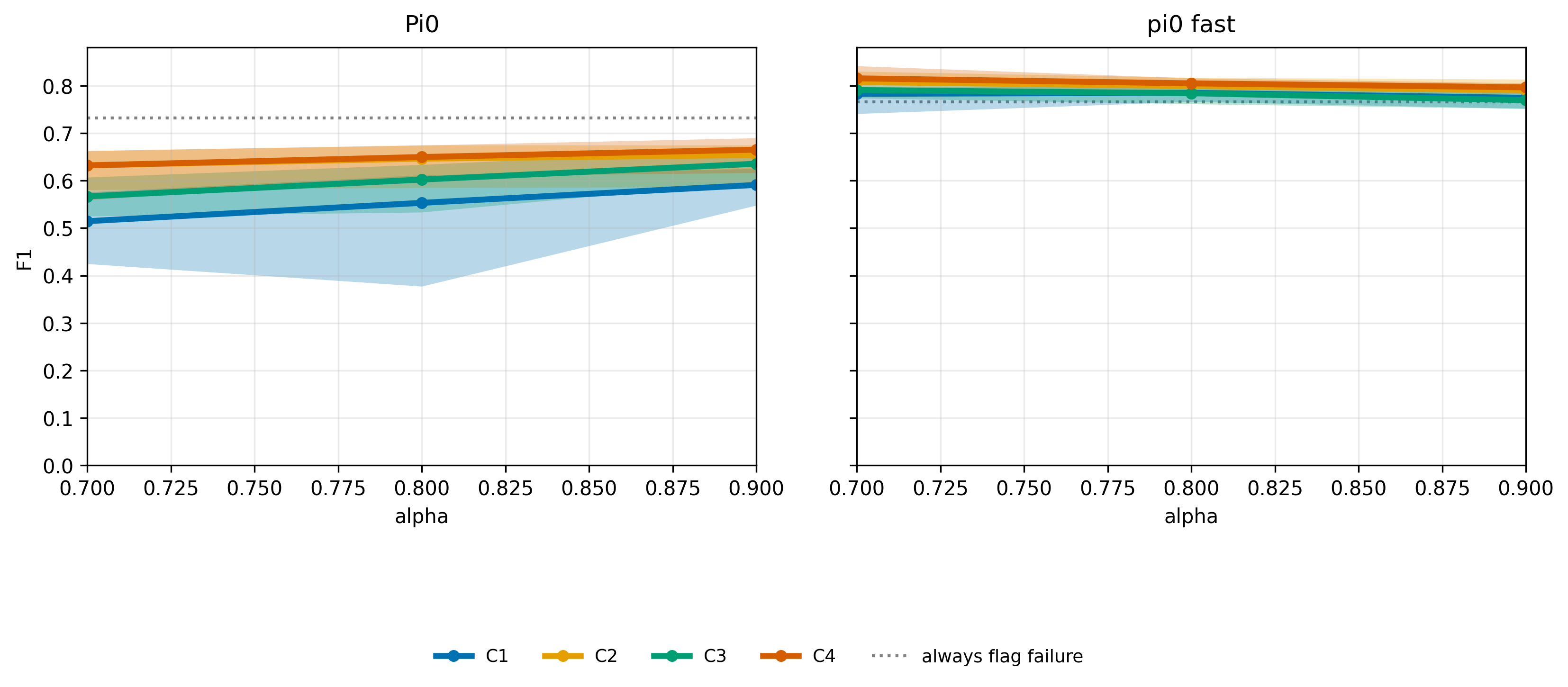}
    \caption{
    Real-world failure detection in the high-$\alpha$ regime, emphasizing conservative intervention when missed failures are costly.
    }
    \label{fig:real_high_alpha}
\end{figure}

Figures~\ref{fig:real_low_alpha} and~\ref{fig:real_high_alpha} show the corresponding real-world DROID trends. 
High-$\alpha$ performance is especially important in real deployment because missed failures can lead to unsafe or unrecoverable robot behavior. 
Together, the low- and high-$\alpha$ plots show whether gains persist across both permissive and conservative operating regimes.

\section{Additional Contrast Set Variations and Per-Shift Results}
\label{app:libero_plus_variations}

We developed and analyzed the effects of an extended set of vision, language, and anti-goal perturbations to augment the single distractor object (vision) and command paraphrase (language) perturbations explored in our primary experiments. 
We find that naively adding perturbations to the SAFECAST training and calibration set does not assist with failure detection at inference time.
Pooling all proposed visual, language, or both visual and language perturbations generally underperforms using only the distractor object and language paraphrase perturbations for training and calibrating the SAFECAST probe for both $\pi_0$ and OpenVLA on both F1 and ROC-AUC metrics of failure detection.
Additionally, we find that multiple family-level variations outperform individual variations for lighting, texture, and paraphrase families.
Finally, we explore anti-goal perturbations where a model is instructed \textit{not} to do the original task, and find that these do not improve failure detection. 
However, these anti-goal variations applied at inference time can reduce the $\pi_0$ success rate from the 90\% range to the 50\% range.
The anti-goal variation may prove a useful sanity check for developing VLA benchmarking and checking for overfitting in general; we will explore this line of thought in upcoming research.


\subsection{Perturbation Family Definitions}
\label{app:perturbation_family_definitions}

We document the additional contrast variations used to construct extended LIBERO contrast-set pools for $\Cfour{}$ ablations, together with per-family F1/ROC-AUC and precision--recall curves for $\pi_0$ and OpenVLA.
We organize these variations into three families: vision variations, language variations, and anti-goal variations.
Vision and language variations are goal-preserving contrast sets: they modify the observation or instruction while keeping the original BDDL goal unchanged.
Anti-goal variations are treated as a separate family; these modify the language instruction to negate the original goal, reflected in a change to the sign of the BDDL goal evaluation.
All rollouts are collected from frozen VLAs under perturbed conditions, filtered with DTW-based active rejection when used for contrast-set training, and evaluated on LIBERO-Plus test rollouts.

The extended contrast variations are grouped as follows:
\begin{itemize}[nosep,noitemsep]
    \item \textbf{Vision variations} preserve the original instruction and BDDL success predicate. This family includes lighting changes and tabletop texture changes. 
    \item \textbf{Language variations} are instruction-level shifts that preserve the original task semantics and BDDL success predicate. This family includes passive-voice instructions and negation-except paraphrases.
    \item \textbf{Anti-goal variations:} goal-reversal tasks that intentionally change the task objective and BDDL success predicate. These are evaluated separately and are not pooled with goal-preserving language variations.
\end{itemize}
We next define each of these families and the variations they comprise.

\begin{figure}[t]
    \centering
    \includegraphics[width=\linewidth]{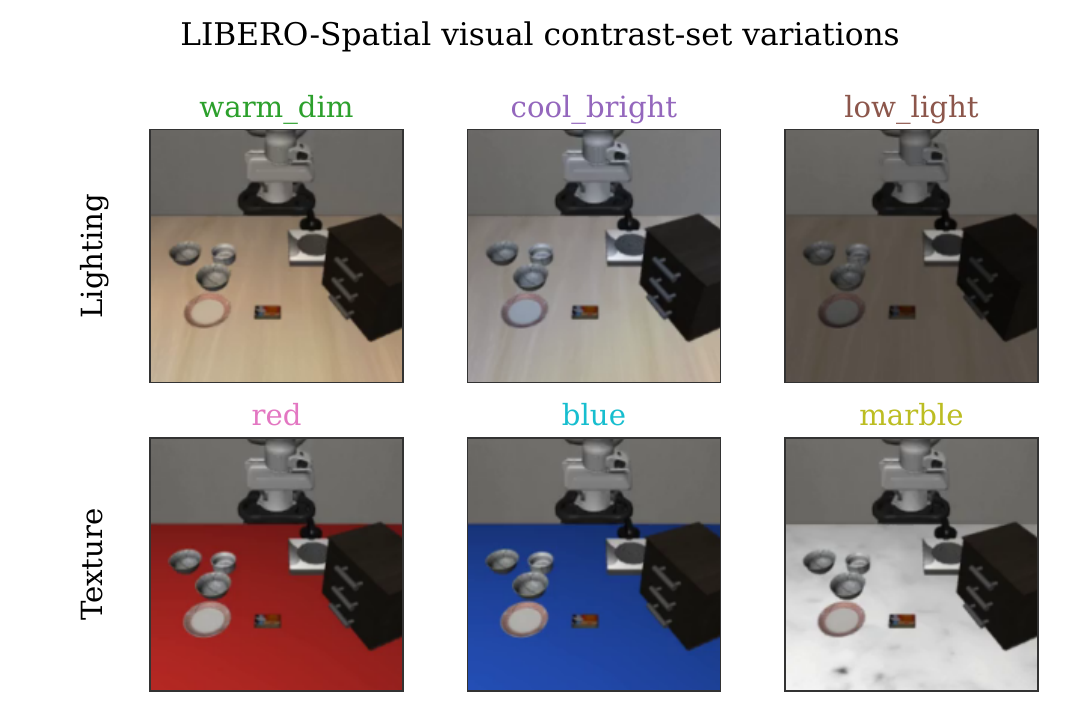}
    \caption{
    Examples of goal-preserving visual contrast variations used in the extended LIBERO contrast-set pool. Each panel shows the first frame of the same task family under a different visual condition, including lighting and tabletop texture changes. These perturbations preserve the instruction, object layout, and BDDL success predicate, so any performance change reflects sensitivity to visual appearance rather than a change in task semantics.
    }
    \label{fig:visual_variations}
\end{figure}

\subsection{Lighting Variations}
\label{app:lighting_variations}

We apply three MuJoCo lighting configurations to the LIBERO-Spatial tabletop scene (Figure~\ref{fig:visual_variations}).
Each configuration modifies diffuse (specular) color, light position, direction, and, for the low light configuration, skybox ambient color.
Table~\ref{tab:lighting_variations} summarizes the settings; rollouts use the same task instructions and success predicates as the source distribution.

\begin{table}[htbp]
\centering
\small
\caption{Lighting contrast-set variations for LIBERO-Spatial. Each lighting condition changes scene illumination while preserving the robot task, object layout, instruction, and BDDL success predicate. These shifts test whether SAFECAST remains robust when the same task is observed under different photometric conditions, such as dim, bright, or color-shifted lighting.}
\label{tab:lighting_variations}
\begin{tabular}{p{0.18\linewidth}p{0.76\linewidth}}
\toprule
Variation & Description \\
\midrule
\texttt{warm\_dim}
& Warm, dimmer front-right lighting with amber diffuse tones and reduced specular highlights. \\
\texttt{cool\_bright}
& Cool blue-tinted, brighter overhead lighting with higher diffuse/specular intensity. \\
\texttt{low\_light}
& Very low scene illumination with dim skybox ambient light, producing near-dark conditions. \\
\texttt{combined\_lighting}
& Pooled training/calibration rollouts from all three single lighting shifts. \\
\bottomrule
\end{tabular}
\end{table}

\subsection{Texture Variations}
\label{app:texture_variations}

We replace the default tabletop appearance with three surface styles.
Object geometry, placement, and task semantics are unchanged.
Table~\ref{tab:texture_variations} lists the variations and representative RGB/texture settings.

\begin{table}[htbp]
\centering
\small
\caption{Table-surface texture contrast-set variations for LIBERO-Spatial. Each texture condition changes the visual appearance of the tabletop while leaving object geometry, object placement, instruction, and task success predicate unchanged. These variations test whether the failure detector relies on brittle background or surface appearance cues rather than task-relevant behavior.}
\label{tab:texture_variations}
\begin{tabular}{p{0.15\linewidth}p{0.22\linewidth}p{0.57\linewidth}}
\toprule
Variation & Mode & Description / example \\
\midrule
\texttt{red}
& Flat color & Saturated red tabletop, RGB $(0.78, 0.18, 0.16)$. \\
\texttt{blue}
& Flat color & Saturated blue tabletop, RGB $(0.16, 0.34, 0.78)$. \\
\texttt{marble}
& Image texture & White marble floor texture mapped to the table surface. \\
\texttt{combined\_texture}
& Pooled & Pooled rollouts from \texttt{red}, \texttt{blue}, and \texttt{marble}. \\
\bottomrule
\end{tabular}
\end{table}

\paragraph{Example.}
For the task \emph{pick up the black bowl on the stove and place it on the plate}, the only change under the \texttt{red} variation is that the robot observes a red tabletop instead of the default wood-like surface; the instruction and goal remain identical.

\subsection{Language Variations}
\label{app:language_variations}

Beyond the LLM-generated semantic paraphrases in Appendix~\ref{app:perturbation_types}, we evaluate two additional goal-preserving language shifts on LIBERO-Spatial: passive voice and \texttt{neg\_except}.
Both perturbations change the surface form or syntactic structure of the instruction while preserving the target object, action, goal location, and BDDL success predicate.
Table~\ref{tab:language_variation_examples} contrasts these perturbations on a representative task.

\begin{table}[htbp]
\centering
\small
\caption{
Goal-preserving language contrast variations with representative instructions. Passive voice and \texttt{neg\_except} change the syntactic form of the command while preserving the target object, action, goal location, and BDDL success predicate. These examples test whether the policy and failure detector remain robust to linguistic reformulations that should imply the same robot trajectory.}
\label{tab:language_variation_examples}
\resizebox{\linewidth}{!}{%
\begin{tabular}{p{0.18\linewidth}p{0.48\linewidth}p{0.28\linewidth}}
\toprule
Variation & Example instruction & Success criterion \\
\midrule
Source (active) &
Pick up the black bowl on the stove and place it on the plate. &
Bowl on plate \\
\hdashline
\texttt{passive} &
The black bowl on the stove is picked up and placed on the plate. &
Bowl on plate; unchanged \\
\hdashline
\texttt{neg\_except} &
Pick up the black bowl on the stove and do not place it anywhere except on the plate. &
Bowl on plate; unchanged \\
\bottomrule
\end{tabular}%
}
\end{table}

\subsubsection{Passive-voice paraphrases}
\label{app:passive_paraphrases}

Passive instructions are generated by converting each stock LIBERO-Spatial BDDL instruction from active to passive voice while preserving the target object, action, and goal location.
We use one passive variant per task.
Given an active instruction of the form
\emph{``Pick up [source description] and place it on the plate,''}
the passive form is
\begin{quote}
\textit{``The [source description] is picked up and placed on the plate.''}
\end{quote}
No new objects, goals, or spatial references are introduced.

\subsubsection{Negation-except paraphrases}
\label{app:neg_except_paraphrases}

\texttt{neg\_except} applies a goal-preserving negation template that forbids incorrect destinations but still requires placing the bowl on the plate.
The template is:
\begin{quote}
\textit{``Pick up [source description] and do not place it anywhere except on the plate.''}
\end{quote}
The BDDL success predicate remains unchanged, e.g.,
\texttt{(On akita\_black\_bowl\_1 plate\_1)}.
This perturbation tests whether the policy can parse negated phrasing without changing the intended task.

\subsection{Anti-Goal Variations}
\label{app:anti_goal_variations}

Anti-goal variations are not language-only perturbations.
They define a separate contrast family in which the task objective itself is reversed along with the language instruction.
This perturbation is the only goal-altering variation considered in our study so far.
Unlike passive voice or \texttt{neg\_except}, which preserve the original task semantics, anti-goal variations change the BDDL success predicate and therefore correspond to a different goal condition.

For each LIBERO-Spatial task, we generate a companion BDDL file using \texttt{scripts/libero\_anti\_goal.py}.
The anti-goal construction modifies three elements:
\begin{enumerate}
    \item \textbf{Language:} the instruction is rewritten to explicitly avoid the original target location.
    \item \textbf{Goal predicate:} the success predicate is changed from the original goal to its negation, e.g., from \texttt{(On bowl plate)} to \texttt{(Not (On bowl plate))}.
    \item \textbf{Initial state:} when needed, the target object is initialized so that satisfying the anti-goal requires behavior different from the source task.
\end{enumerate}

\begin{table}[htbp]
\centering
\small
\caption{
Anti-goal contrast variation. Unlike goal-preserving language perturbations, anti-goal changes the task objective and therefore changes the BDDL success predicate. This setting tests whether the failure detector behaves differently under explicit goal reversal, where the instruction no longer describes the original source task.
}
\label{tab:anti_goal_example}
\resizebox{\linewidth}{!}{%
\begin{tabular}{p{0.20\linewidth}p{0.48\linewidth}p{0.26\linewidth}}
\toprule
Setting & Example instruction & Success criterion \\
\midrule
Source task &
Pick up the black bowl on the stove and place it on the plate. &
Bowl on plate \\
\hdashline
Anti-goal task &
Pick up the black bowl and place it anywhere except on the plate. &
Bowl not on plate \\
\bottomrule
\end{tabular}%
}
\end{table}

Because anti-goal changes the success predicate, it is evaluated as a separate family and is not included in \texttt{combined\_language}.
This separation avoids conflating semantics-preserving language robustness with robustness to explicit goal reversal.

\subsection{Active Rejection During Contrast-Set Collection}
\label{app:active_rejection_variations}

All contrast-set rollouts in this section, lighting, texture, passive, \texttt{neg\_except}, are collected with DTW-based active rejection when used for probe training.
During collection, each candidate rollout is represented by a translation-normalized end-effector trajectory; FastDTW distance is normalized by warping-path length and compared against already-accepted rollouts in the same task-and-outcome pool.
Candidates with normalized DTW distance below $\tau_{\mathrm{DTW}} = 0.02$ are rejected.
This reduces near-duplicate trajectories in $\DcsTrain$ and increases diversity among accepted perturbation rollouts before probe training and conformal calibration.

\subsection{\Cfour\ Pooling Configurations}
\label{app:c4_pooling}

Individual perturbations above are pooled into broader $\Cfour{}$ training/calibration subsets:
\begin{itemize}
    \item \textbf{Pooled lighting} (\texttt{c4\_combined\_lighting}): \texttt{warm\_dim} + \texttt{cool\_bright} + \texttt{low\_light}.
    \item \textbf{Pooled texture} (\texttt{c4\_combined\_texture}): \texttt{red} + \texttt{blue} + \texttt{marble}.
    \item \textbf{Vision} (\texttt{c4\_combined\_vision}): lighting + texture.
    \item \textbf{Language} (\texttt{c4\_combined\_language}): passive + \texttt{neg\_except}.
    \item \textbf{Vision + language} (\texttt{c4\_combined\_vision\_language}): vision variations plus passive and \texttt{neg\_except}.
    \item \textbf{Anti-goal} (\texttt{c4\_anti\_goal}): goal-reversal tasks only.
\end{itemize}

Each $\Cfour{}$ ablation trains and calibrates the probe on source rollouts plus the listed contrast pool, then evaluates on LIBERO-Plus unseen rollouts.
When additional perturbation pools are added, the available training/calibration pool becomes larger; however, batch size and number of training iterations are kept fixed so that each probe receives the same number of gradient updates.
Anti-goal is always kept separate from the language pool because it changes the task success predicate.

\subsection{F1 and Cumulative ROC-AUC Curves}
\label{app:variation_f1_roc_quads}

\begin{figure*}[t]
    \centering
    \includegraphics[width=\linewidth]{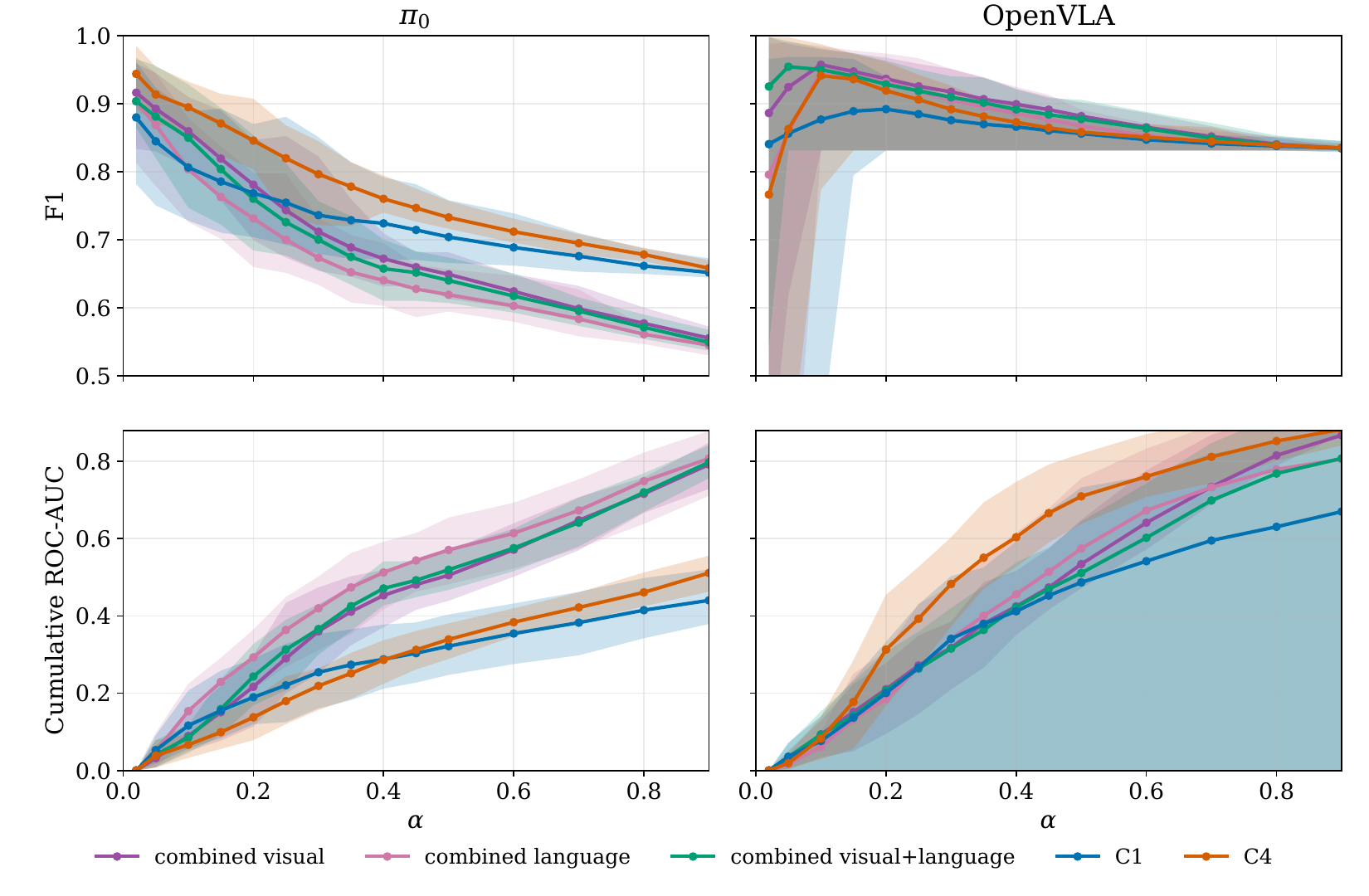}
    \caption{Combined contrast-family ablations for $\Cfour{}$ on LIBERO-Plus. The top row reports F1 versus conformal risk level $\alpha$, where higher curves indicate better calibrated failure detection at that operating point. The bottom row reports cumulative ROC-AUC versus $\alpha$, where higher curves indicate stronger separation between successful and failed rollouts across the conformal threshold sweep. The left column shows $\pi_0$ and the right column shows OpenVLA. For $\pi_0$, the original C4 reference maintains the strongest F1 across most of the sweep, while the combined vision-language pool improves ROC-AUC at higher $\alpha$, suggesting better separability but a precision--recall trade-off. For OpenVLA, C4 and the combined vision-language setting are strongest overall, indicating that OpenVLA benefits most when visual and language shifts are modeled jointly. Overall, joint vision-language contrast sets provide the most consistent robustness signal, but their benefit is clearer in ROC-AUC than in F1 for $\pi_0$.}
    \label{fig:c4_combined_quad}
\end{figure*}

\begin{figure*}[t]
    \centering
    \includegraphics[width=\linewidth]{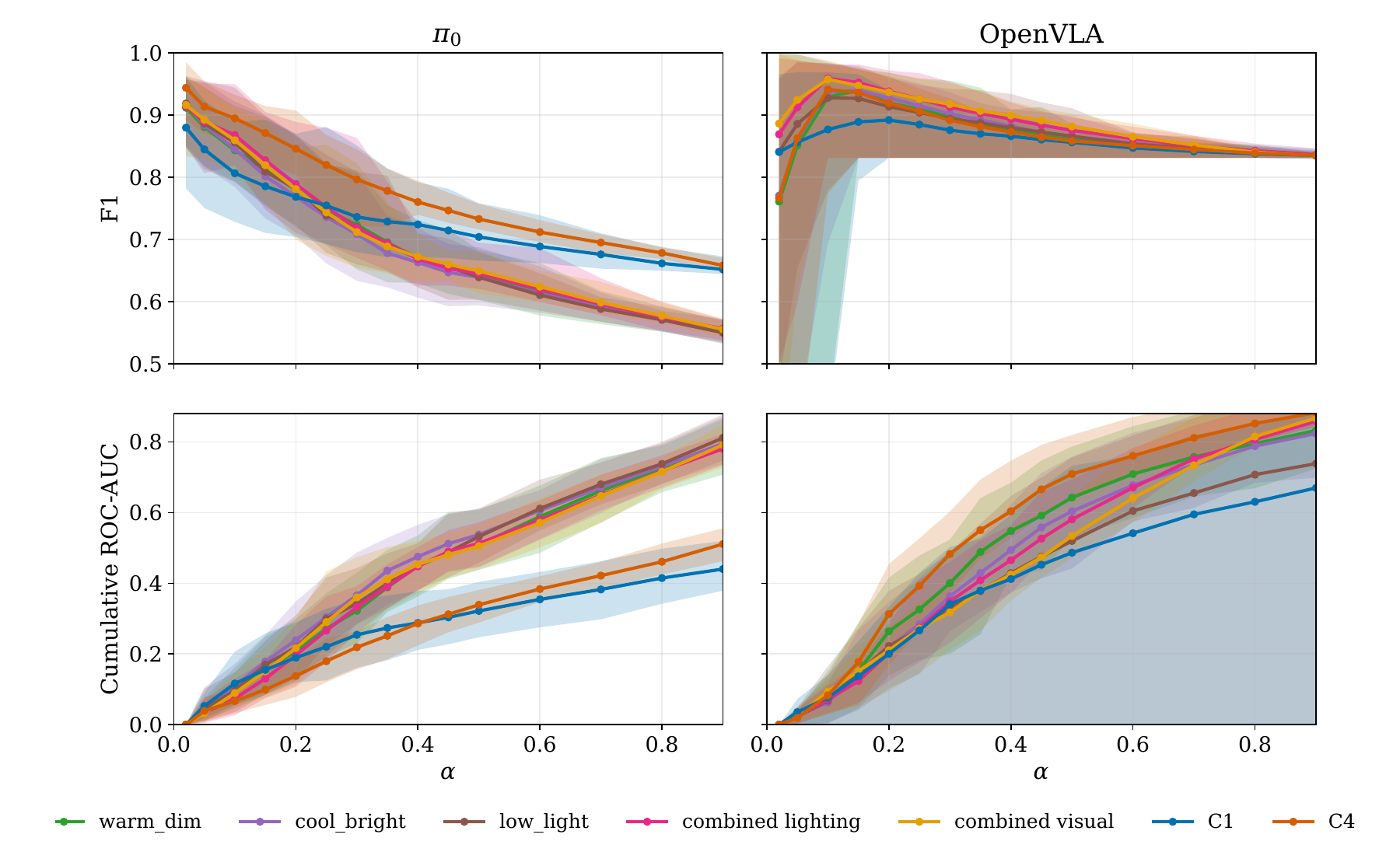}
    \caption{ Lighting variation ablations for $\Cfour{}$ on LIBERO-Plus. The curves compare individual lighting shifts, pooled lighting, the combined-vision pool, and the C1/C4 references. Higher F1 in the top row indicates better calibrated failure detection at a fixed conformal risk level $\alpha$, while higher cumulative ROC-AUC in the bottom row indicates better success/failure separability as the conformal threshold sweep expands. For $\pi_0$, lighting-only variants improve separability over C1 at higher $\alpha$, but they do not consistently exceed the original C4 reference in F1, suggesting that photometric diversity alone is not sufficient for the best calibrated detector. For OpenVLA, the C4 and combined-vision curves remain among the strongest, indicating that lighting changes are useful but work best when combined with other visual perturbations. Thus, lighting perturbations improve coverage of visual shift, but pooled visual diversity is generally more reliable than any single lighting condition.}
    \label{fig:c4_lighting_quad}
\end{figure*}

\begin{figure*}[t]
    \centering
    \includegraphics[width=\linewidth]{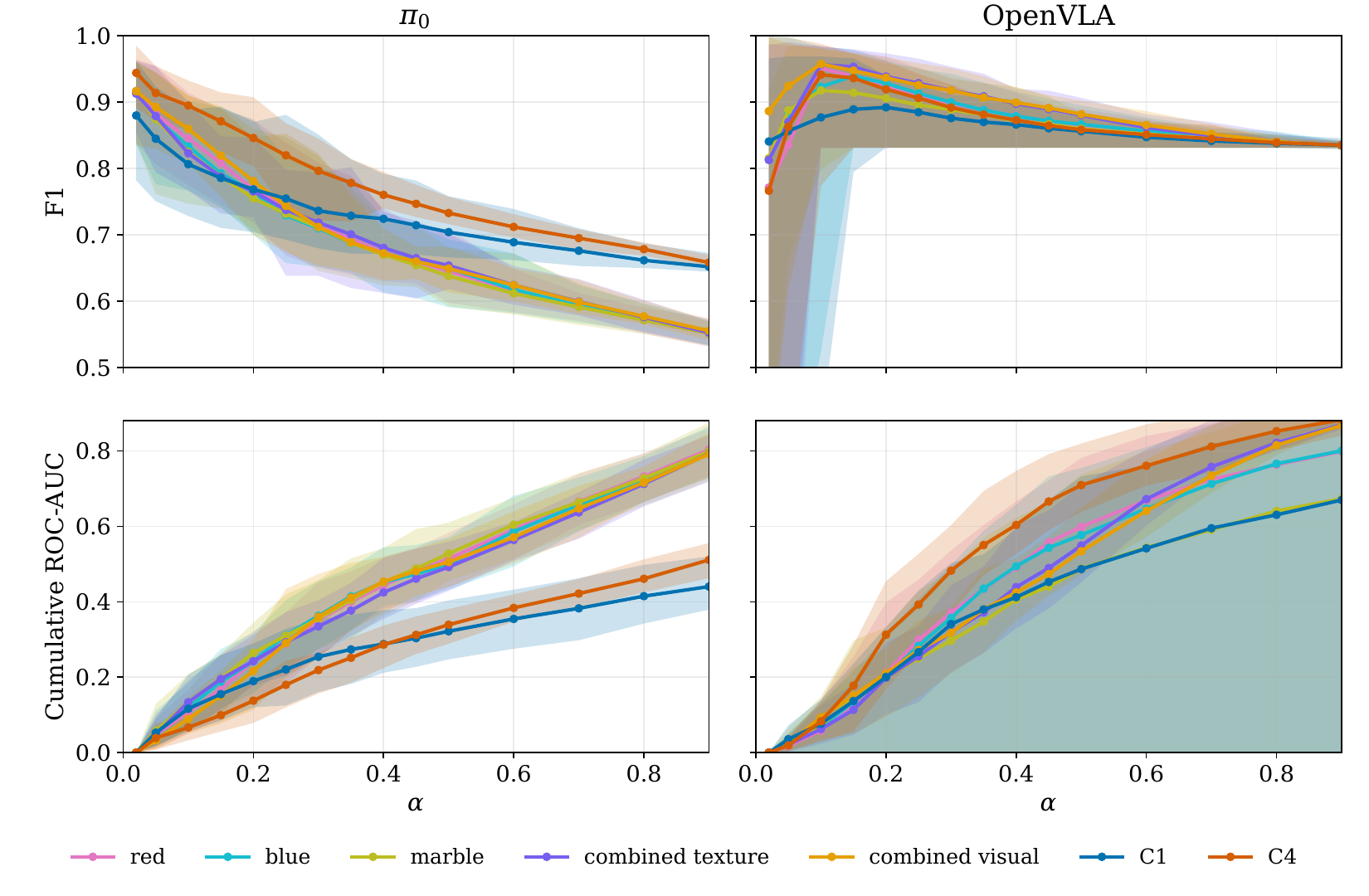}
    \caption{
    Texture variation ablations for $\Cfour{}$ on LIBERO-Plus. The curves compare red, blue, and marble tabletop textures, pooled texture, the combined-vision pool, and the C1/C4 references. Because these perturbations preserve the instruction, object layout, and success predicate, performance differences measure sensitivity to background and surface appearance rather than changes in task semantics. For $\pi_0$, texture variants can improve cumulative ROC-AUC relative to C1 at larger $\alpha$, showing that surface-appearance diversity helps separate successful and failed rollouts, but F1 remains strongest or most stable for the C4 reference. For OpenVLA, C4 and combined-vision remain the most reliable overall, with texture-only shifts providing smaller gains. This suggests that texture changes are useful as part of a broader visual contrast pool, but are weaker than the full visual contrast setting by themselves.
    }
    \label{fig:c4_texture_quad}
\end{figure*}

\begin{figure*}[t]
    \centering
    \includegraphics[width=\linewidth]{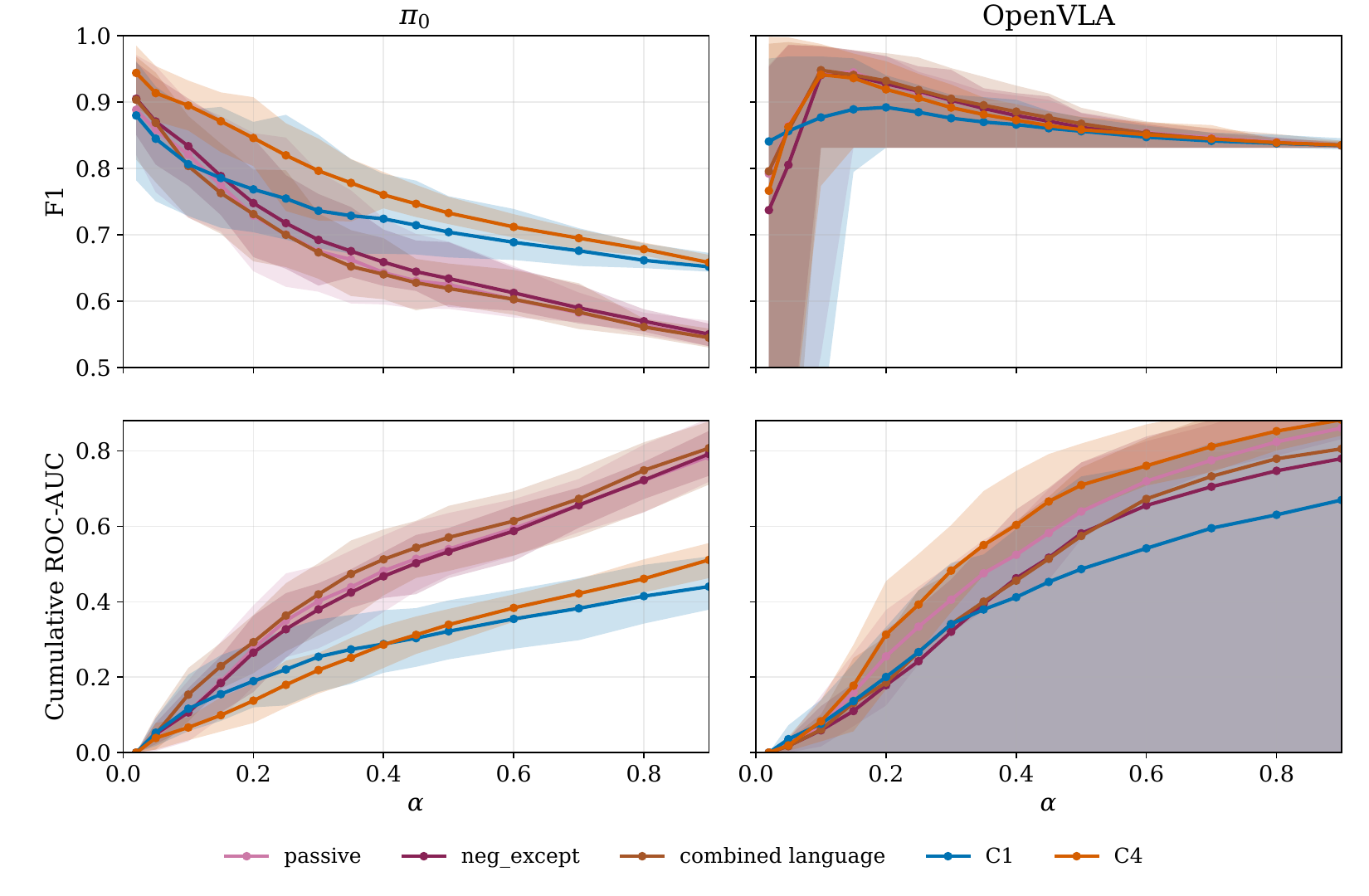}
    \caption{Goal-preserving language variation ablations for $\Cfour{}$ on LIBERO-Plus. The curves compare passive voice, \texttt{neg\_except}, pooled language, and the C1/C4 references. All language variants preserve the original BDDL success predicate, so changes in F1 and ROC-AUC reflect robustness to instruction form rather than a change in the desired goal. For $\pi_0$, passive and \texttt{neg\_except} improve ROC-AUC over C1 at higher $\alpha$, indicating that syntactic language diversity can improve failure separability, but their F1 curves are lower than or comparable to the C4 reference across much of the sweep. For OpenVLA, language variants closely track the stronger C4 behavior, suggesting that OpenVLA is especially sensitive to, and benefits from, language-side contrast information. Overall, goal-preserving language perturbations are useful, but the strongest robustness comes from combining language with visual contrast sets rather than using language alone.}
    
    \label{fig:c4_language_quad}
\end{figure*}

\begin{figure*}[t]
    \centering
    \includegraphics[width=\linewidth]{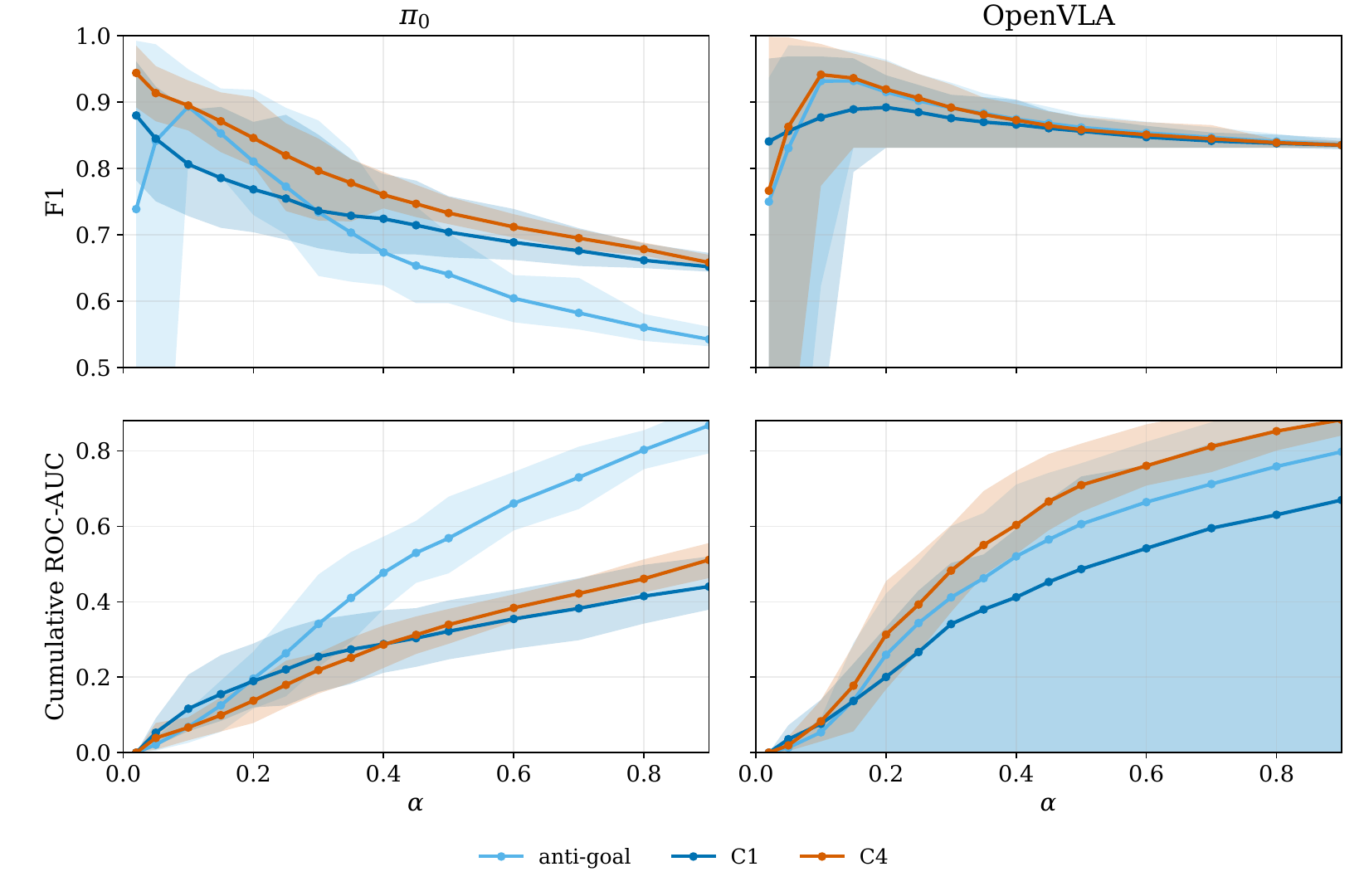}
    \caption{
    Anti-goal ablation for $\Cfour{}$ on LIBERO-Plus. Anti-goal is evaluated separately from language perturbations because it reverses the task objective and changes the BDDL success predicate. The top row shows F1 across $\alpha$, and the bottom row shows cumulative ROC-AUC; higher curves are better for both metrics. For $\pi_0$, anti-goal produces high ROC-AUC at larger $\alpha$ but weaker F1 than C4, indicating that goal reversal can make successes and failures more separable while also changing the calibrated precision--recall trade-off. For OpenVLA, C4 remains stronger overall, especially in ROC-AUC, suggesting that ordinary contrast-set training and calibration are more useful than anti-goal exposure for deployment-style LIBERO-Plus shifts. This result supports treating anti-goal as a distinct stress test rather than pooling it with goal-preserving language perturbations.
    }
    \label{fig:c4_anti_goal_quad}
\end{figure*}

Figures~\ref{fig:c4_combined_quad}--\ref{fig:c4_anti_goal_quad} report seed-averaged F1 and cumulative ROC-AUC versus conformal risk level $\alpha$ for $\pi_0$ and OpenVLA.
Each figure uses a $2\times2$ layout: F1 is shown in the top row, cumulative ROC-AUC in the bottom row, $\pi_0$ in the left column, and OpenVLA in the right column.
Shaded bands denote p10--p90 across 30 random seeds.
F1 axes are fixed to $[0.5, 1.0]$ for readability.
Curves labeled C1 and C4 provide source-only and full contrast-set references from the main paper.

\paragraph{How to read the families.}
\begin{itemize}
    \item \textbf{Combined} (Fig.~\ref{fig:c4_combined_quad}): compares vision, language, and joint vision+language $\Cfour{}$ configurations.
    \item \textbf{Lighting} (Fig.~\ref{fig:c4_lighting_quad}): compares single lighting shifts, \texttt{combined\_lighting}, and the \texttt{combined\_vision} reference.
    \item \textbf{Texture} (Fig.~\ref{fig:c4_texture_quad}): compares single texture shifts, \texttt{combined\_texture}, and the \texttt{combined\_vision} reference.
    \item \textbf{Language} (Fig.~\ref{fig:c4_language_quad}): compares passive, \texttt{neg\_except}, and \texttt{combined\_language}.
    \item \textbf{Anti-goal} (Fig.~\ref{fig:c4_anti_goal_quad}): evaluates goal-reversal contrast only and tests failure detection under instruction--goal mismatch.
\end{itemize}

\subsection{LIBERO-Plus Evaluation Class Balance}
\label{app:liberoplus_class_balance}

Table~\ref{tab:liberoplus_test_sr} reports the base policy success rates on the LIBERO-Plus test rollouts used for SAFECAST evaluation.
These success rates determine the success/failure class balance for failure detection: $\pi_0$ succeeds on a majority of LIBERO-Plus rollouts, while OpenVLA fails on a majority of rollouts.
This difference is important when interpreting ROC-AUC and precision--recall curves, because the two policies induce different failure prevalences on the same shifted evaluation suite.

\begin{table}[htbp]
\centering
\small
\caption{Policy success rate (SR) on the LIBERO-Plus test set used for SAFECAST evaluation.}
\label{tab:liberoplus_test_sr}
\begin{tabular}{lccc}
\toprule
Policy & Successes & Rollouts & SR \\
\midrule
$\pi_0$ & 1{,}546 & 2{,}406 & 64.3\% \\
OpenVLA & 693 & 2{,}406 & 28.8\% \\
\bottomrule
\end{tabular}
\end{table}

Because failure detection treats failed rollouts as the positive class, the implied failure rates are 35.7\% for $\pi_0$ and 71.1\% for OpenVLA.
This difference partly motivates the precision--recall diagnostics in Appendix~\ref{app:pr_curves}, since PR curves are more sensitive than ROC curves to the positive-class prevalence.

\subsection{Precision--Recall Curves}
\label{app:pr_curves}

We additionally report precision--recall diagnostics for the extended contrast-set perturbations.
These curves are included because some perturbation families, especially for $\pi_0$, produce unusual ROC-AUC trends that may be affected by class imbalance or threshold behavior.
Precision--recall curves provide a complementary view of whether increased failure recall is accompanied by useful precision.

For each conformal risk level $\alpha$ in the functional CP sweep, we extract F1 and true-positive rate (TPR) on LIBERO-Plus test rollouts and derive precision as
\[
\mathrm{Prec}(\alpha) =
\frac{\mathrm{F1}(\alpha)\cdot \mathrm{TPR}(\alpha)}
{2\,\mathrm{TPR}(\alpha)-\mathrm{F1}(\alpha)}.
\]
Here, TPR is equivalent to recall for the failure-detection task.
When $2\mathrm{TPR}(\alpha)-\mathrm{F1}(\alpha)=0$, precision is undefined and the corresponding point is omitted.
Points are connected in $\alpha$ order.
Thus, these curves summarize how calibrated failure detection trades precision and recall as the conformal threshold varies.

\begin{figure*}[t]
    \centering
    \includegraphics[width=\linewidth]{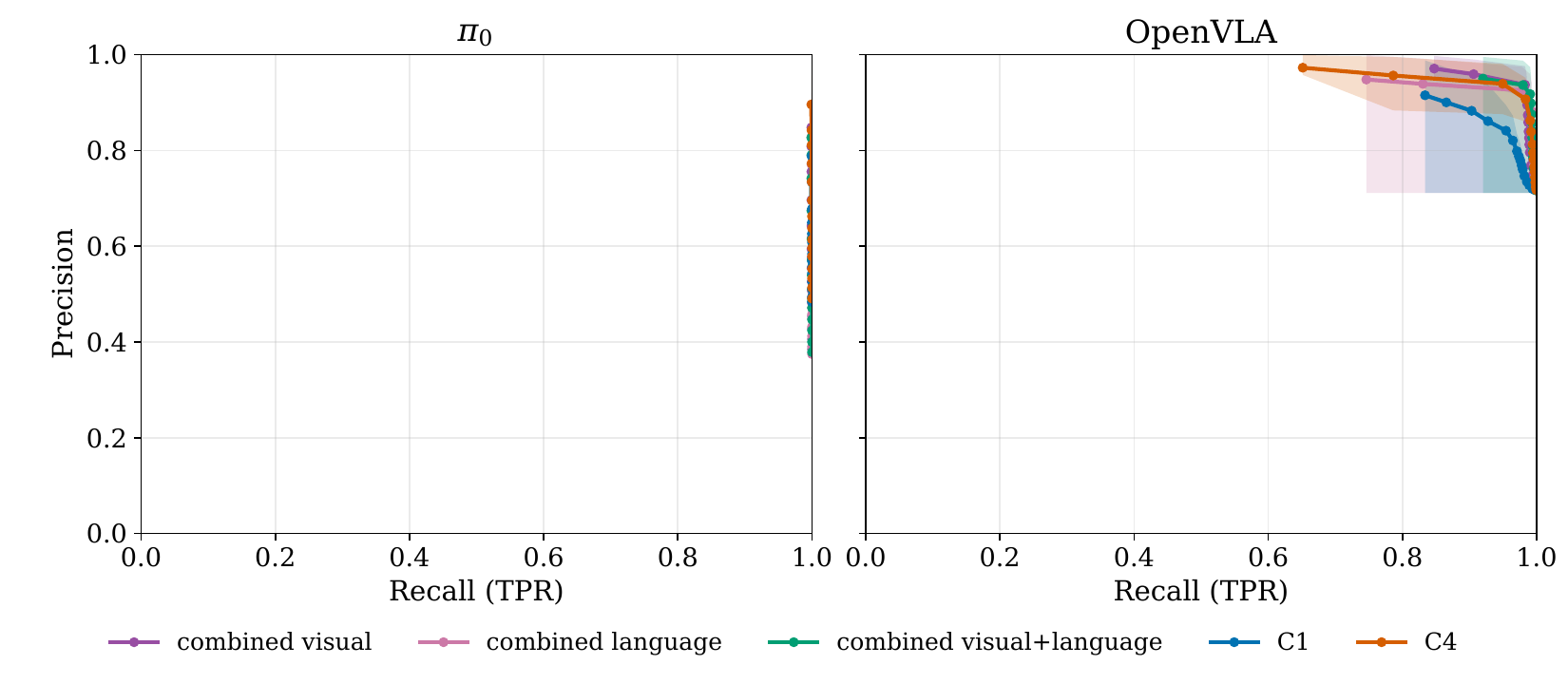}
    \caption{
    Precision--recall curves for combined $\Cfour{}$ ablations on LIBERO-Plus. Each point corresponds to one conformal risk level $\alpha$; curves farther toward the upper-right are better because they achieve higher precision at higher recall. The left panel shows $\pi_0$ and the right panel shows OpenVLA. For $\pi_0$, most curves operate near very high recall, so differences are mainly visible in precision; this indicates that several settings flag most failures, but differ in how many false alarms they introduce. For OpenVLA, the C4 and joint vision-language curves lie closest to the upper-right envelope, showing the best balance between catching failures and avoiding unnecessary interventions. This PR view clarifies that ROC-AUC gains should be interpreted together with precision, especially for $\pi_0$ where class balance makes many methods appear recall-heavy.
    }
    \label{fig:c4_pr_combined}
\end{figure*}

\begin{figure*}[t]
    \centering
    \includegraphics[width=\linewidth]{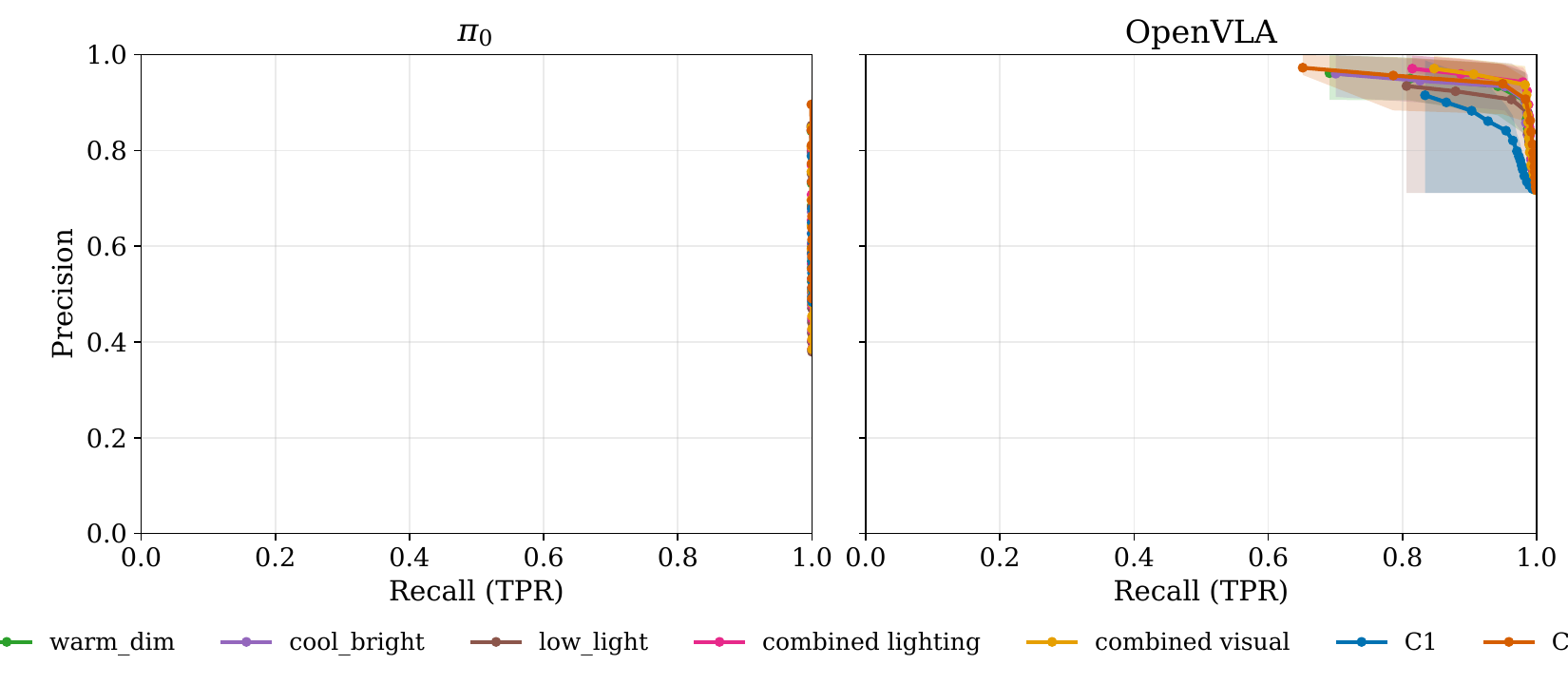}
    \caption{
    precision--recall curves for lighting variation ablations on LIBERO-Plus. Curves closer to the upper-right indicate better calibrated failure detection, with higher recall meaning more failures are caught and higher precision meaning fewer false alarms. For $\pi_0$, lighting variants largely cluster at high recall, so the key distinction is whether they preserve precision while flagging many failures. For OpenVLA, C4 and the stronger pooled visual settings maintain the best precision--recall trade-off, while individual lighting shifts are more variable. This suggests that lighting perturbations help expose photometric failure modes, but lighting alone is less reliable than broader visual contrast diversity.
    }
    \label{fig:c4_pr_lighting}
\end{figure*}

\begin{figure*}[t]
    \centering
    \includegraphics[width=\linewidth]{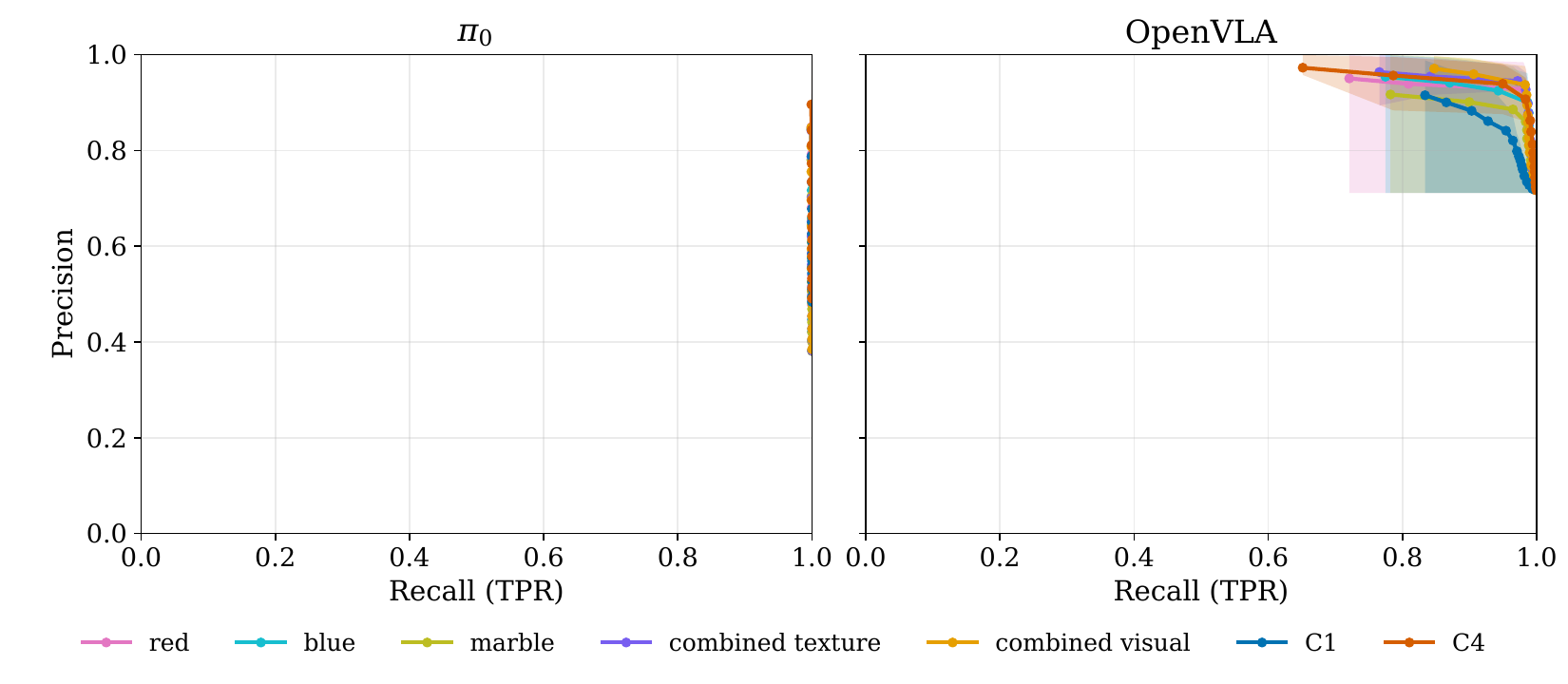}
    \caption{
    Precision--recall curves for texture variation ablations on LIBERO-Plus. The curves compare individual tabletop texture shifts, pooled texture, combined vision, and the C1/C4 references. Higher and farther-right curves are better because they detect more failed rollouts while maintaining higher precision. For $\pi_0$, texture variants again concentrate near high recall, indicating aggressive failure coverage, but precision is the deciding factor for practical usefulness. For OpenVLA, the C4 and combined-vision references remain closest to the upper-right region, suggesting that texture changes are helpful but work best as part of a larger visual perturbation pool. These curves support the interpretation from the F1/ROC-AUC plots: texture diversity improves visual robustness, but by itself is not consistently the best contrast-set family.
    }
    \label{fig:c4_pr_texture}
\end{figure*}

\begin{figure*}[t]
    \centering
    \includegraphics[width=\linewidth]{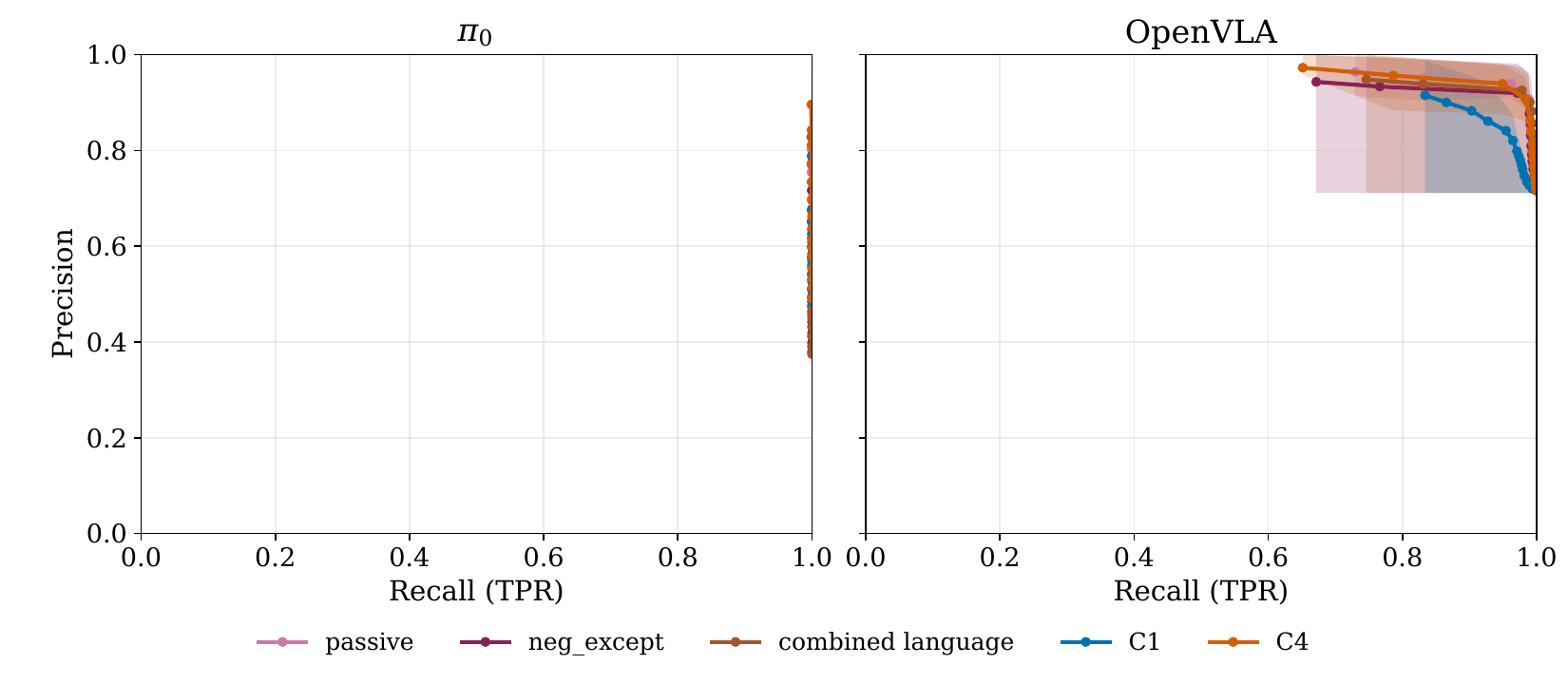}
    \caption{
    Precision--recall curves for goal-preserving language variation ablations on LIBERO-Plus. The curves compare passive voice, \texttt{neg\_except}, pooled language, and the C1/C4 references. Because all variants preserve the original task goal, upper-right movement in this plot indicates improved robustness to instruction phrasing rather than to a changed task objective. For $\pi_0$, the language variants primarily affect precision at very high recall, showing that they tend to detect many failures but may differ in false-alarm behavior. For OpenVLA, language contrast sets are competitive with C4, indicating that language-side perturbations provide a strong robustness signal for this policy. Overall, these PR curves show that passive and negation-based instructions help evaluate whether failure detection remains useful under syntactic instruction shifts..
    }
    \label{fig:c4_pr_language}
\end{figure*}

\begin{figure*}[t]
    \centering
    \includegraphics[width=\linewidth]{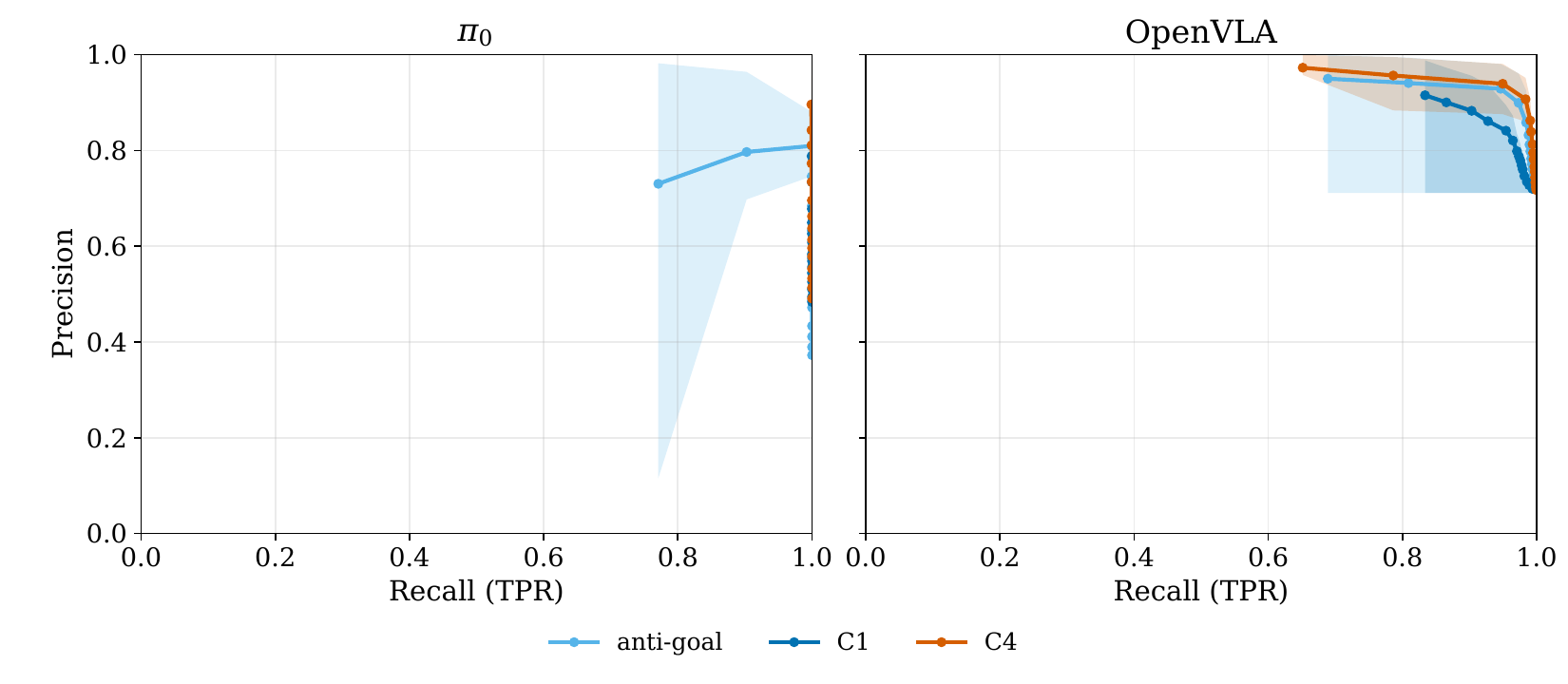}
    \caption{
    Pecision--recall curves for anti-goal contrasts on LIBERO-Plus. Anti-goal reverses the task objective and is therefore shown separately from goal-preserving language perturbations. Curves closer to the upper-right indicate better practical failure detection, combining high failure recall with high precision. For $\pi_0$, the anti-goal curve reaches high recall but has more variable precision, consistent with the F1/ROC-AUC result that anti-goal improves separability without necessarily improving calibrated detection quality. For OpenVLA, C4 maintains a stronger precision--recall trade-off than anti-goal, suggesting that goal-reversal data is less aligned with the LIBERO-Plus deployment shifts than the original contrast-set construction. This confirms that anti-goal is best interpreted as a severe stress test rather than a replacement for goal-preserving language contrast sets.
    }
    \label{fig:c4_pr_anti_goal}
\end{figure*}

Figures~\ref{fig:c4_pr_combined}--\ref{fig:c4_pr_anti_goal} report CP-$\alpha$ precision--recall curves for the combined, lighting, texture, language, and anti-goal perturbation families.

The precision--recall diagnostics are especially important for interpreting the $\pi_0$ results.
Because $\pi_0$ succeeds on many source-distribution and mildly shifted rollouts, some perturbation settings can produce class imbalance in the evaluation labels.
In such cases, ROC-AUC may improve even when calibrated failure detection does not uniformly improve across operating points.
The CP-$\alpha$ precision--recall curves decompose this behavior by showing whether a perturbation mainly increases recall, preserves precision, or introduces a precision--recall trade-off.

\subsection{Discussion and Key Takeaways}

We developed and analyzed the effects of an extended set of vision, language, and anti-goal perturbations to augment the single distractor object (vision) and command paraphrase (language) perturbations explored in our primary experiments. 

\paragraph{Naively Adding Perturbations Does Not Help Failure Detection.}
We find that pooling all proposed visual, language, or both visual and language perturbations generally underperforms using only the distractor object and language paraphrase perturbations for training and calibrating the SAFECAST probe for both $\pi_0$ and OpenVLA on both F1 and ROC-AUC metrics of failure detection (Figure~\ref{fig:c4_combined_quad}).
A notable exception is the cumulative ROC-AUC of the $\pi_0$ model, which is consistently improved by these expanded contrast sets across sensitivity values $\alpha$. 
A partial accounting of this effect comes from the class imbalance in $\pi_0$ failure detection on the LIBERO-Plus test set we utilize; $\pi_0$ achieves success on most test instances, making sensitivity to correctly identifying failure more visibly rewarded in ROC-AUC than F1, which is more easily flattened by majority class predictions. 
See the precision and recall curves presented in Figures~\ref{fig:c4_pr_combined} through~\ref{fig:c4_pr_anti_goal} for $\pi_0$ and OpenVLA for a visual accounting of this phenomena; in particular, $\pi_0$ SAFE and SAFECAST probes trained across nearly every evaluated contrast set perturbation result in 1.0 recall regardless of $\alpha$ sensitivity. 

\paragraph{Multiple Family-level Variations Outperform Individual Variations.}
In the cases of lighting variations (Figure~\ref{fig:c4_lighting_quad}), tabletop texture variations (Figure~\ref{fig:c4_texture_quad}), and passive voice and negation-except paraphrases (Figure~\ref{fig:c4_language_quad}), we find that combining these perturbations outperforms or matches the performance of each lighting/texture/paraphrase method individually in F1 and ROC-AUC on failure detection. 
Similarly to the findings of naive perturbation combinations, the primary SAFECAST method utilizing only distractor objects and language instruction paraphrases outperforms these combined perturbations in most cases, with $\pi_0$ ROC-AUC again providing a counter-trend that we attribute partially to class imbalance. 

\paragraph{Anti-goal Perturbations Do Not Improve Failure Detection} but they are really, really funny.
Figure~\ref{fig:c4_anti_goal_quad} reveals the same trends for anti-goal as other additional tested vision and language paraphrases: it is outperformed by vanilla SAFECAST except for $\pi_0$ ROC-AUC, where class imbalance lets anti-goal shine by achieving a better precision-recall tradeoff (Figure~\ref{fig:c4_pr_anti_goal}). 
However, a more interesting finding and takeaway for later work is that anti-goal variations applied at inference time to generate probe calibration data reduced the $\pi_0$ success rate from the 90\% range to the 50\% range, effectively rendering its success a coin toss as to whether the system would ignore the command it was told to ignore, or skip over the negation term and fall into an overfit trajectory execution anyway. 
The anti-goal variation may prove a useful sanity check for developing VLA benchmarking and checking for overfitting in general; we will explore this line of thought in upcoming research. 


\end{document}